\documentclass[sn-basic]{sn-jnl}

\jyear{2021}%

\theoremstyle{thmstyleone}%
\theoremstyle{thmstyletwo}%

\theoremstyle{thmstylethree}%

\usepackage{xurl}
\usepackage{url}

\begin{document}

\title[Dermoscopic Dark Corner Artifacts Removal: Friend or Foe?]{Dermoscopic Dark Corner Artifacts Removal: Friend or Foe?}


\author[1]{\fnm{Samuel William} \sur{Pewton}}\email{sam.pewton@hotmail.co.uk}
\equalcont{These authors contributed equally to this work.}

\author[1]{\fnm{Bill} \sur{Cassidy}}\email{b.cassidy@mmu.ac.uk}
\equalcont{These authors contributed equally to this work.}

\author[1]{\fnm{Connah} \sur{Kendrick}}\email{connah.kendrick@mmu.ac.uk}

\author*[1]{\fnm{Moi Hoon} \sur{Yap}}\email{m.yap@mmu.ac.uk}

\affil*[1]{\orgdiv{Faculty of Science and Engineering}, \orgname{Manchester Metropolitan University}, \orgaddress{\street{Chester Street}, \city{Manchester}, \postcode{M1 5GD}, \country{UK}}}




\abstract{One of the more significant obstacles in classification of skin cancer is the presence of artifacts. This paper investigates the effect of dark corner artifacts, which result from the use of dermoscopes, on the performance of a deep learning binary classification task. Previous research attempted to remove and inpaint dark corner artifacts, with the intention of creating an ideal condition for models. However, such research has been shown to be inconclusive due to lack of available datasets labelled with dark corner artifacts and detailed analysis and discussion. To address these issues, we label 10,250 skin lesion images from publicly available datasets and introduce a balanced dataset with an equal number of melanoma and non-melanoma cases. The training set comprises 6126 images without artifacts, and the testing set comprises 4124 images with dark corner artifacts. We conduct three experiments to provide new understanding on the effects of dark corner artifacts, including inpainted and synthetically generated examples, on a deep learning method. Our results suggest that introducing synthetic dark corner artifacts which have been superimposed onto the training set improved model performance, particularly in terms of the true negative rate. This indicates that deep learning learnt to ignore dark corner artifacts, rather than treating it as melanoma, when dark corner artifacts were introduced into the training set. Further, we propose a new approach to quantifying heatmaps indicating network focus using a root mean square measure of the brightness intensity in the different regions of the heatmaps. This paper provides a new guideline for skin lesions analysis with an emphasis on reproducibility.}




\maketitle

\section{Introduction}\label{sec1}

Using microscopy for clinical examination is not a new concept.
The first recorded example of using microscopy dates back to 1655 where Pierre Borel observed capillaries of the nailbed under a microscope. Ever since this moment there have been many studies resulting in improvements to the process, including the use of different immersion fluids to make the upper layers of the epidermis more translucent to improve examination. Portable devices were not available until 1990 where Kreusch and Rassner developed a portable stereomicroscope capable of magnification from 10-40x (\cite{buch2021dermoscopy}). The downside to this device is that it was much more expensive than the devices used previously. These early advancements led to the development of the hand-held dermatoscope (\cite{buch2021dermoscopy}). The use of a dermatoscope gives the dermatologist the ability to magnify and view features of the lesion that were obscure or invisible to the naked eye allowing for a more accurate diagnosis (\cite{hayesdermoscopy}).

Dark corner artifacts (DCA) in skin lesion images can be defined as regions around the edges of the image which are dark in appearance which can vary in intensity. This phenomenon is a result of the circular shape of the dermoscopic lens when pressed against the skin during a dermatological examination. DCA are not always present in dermatological images. Variations in dermatoscope calibration, camera zoom levels, and post-processing, as manually determined by the examining dermatologist, are the main contributing factors in the presence and degree of DCA (\cite{ain2020genetic}). 

\begin{figure}[!htbp]
	\centering
	\begin{tabular}{cc}
	
	    Contact & Non Contact \\
		\includegraphics[width=4cm,height=4cm]{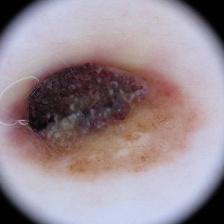} &
		\includegraphics[width=4cm,height=4cm]{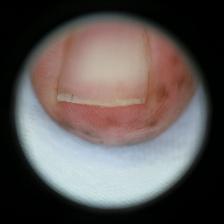} \\
		
	\end{tabular}
	\caption[]{Illustration of Contact vs Non Contract border artifact.}
	\label{fig:DCA}
\end{figure}

Another type of DCA appears when a non-contact dermatoscope is used to take an image of a lesion located on a non-flat surface of skin, such as the ear. In this example, the non-ear region is black due to the focus, exposure and white balance settings of the camera. A card is often placed around the ear (and similar areas of the body such as nose and digits) if the exposure or focus is incorrect due to the camera's attempt to balance all parts of the image. Figure~\ref{fig:DCA} shows examples of both types of artifacts. Additionally, if the nevi of the lesion is large, the dermatologist may not be able to capture the entire lesion when zoomed-in to remove the DCA from the image - this happens regardless of the model of the dermatoscope.

The cost of dermatoscopes may be linked to the number of features available on the device, however, cost is not directly associated with the presence of DCA. The most expensive devices may exhibit DCA, while cheaper ones may not, and vice versa. An example of this is the DermLite HÜD$^{\circledR}$, which is one of the cheapest dermatoscopes. This device provides a rectangular view of the lesion, and is therefore easy to remove DCA. Zoom settings used to remove DCA are a feature only of the camera used to acquire the skin lesion photograph, and not the dermatoscope itself. Cropping of DCA is common practice in dermascopy, which may include cropping of small regions of the dermoscopic view.
Some of the publicly available datasets (\cite{tschandl2018ham}) perform pre-possessing to normalise colours and provide zoomed images to reduce the presence of the DCA when possible as a form of natural augmentation.


Dermoscopic images found in the ISIC datasets are sourced from a variety of dermoscopes and cameras. They may include transparency slide scans, dedicated dermoscopy cameras, video stills, fixed focus devices, lenses of differing diameter (10mm to 30mm), shape and quality. Some are plastic while others are glass, may be scratched or dirty, and may have been acquired with or without immersion contact fluid. Most ISIC images were acquired using direct contact with skin and lesion, with rare cases showing non-contact.

The use of dermoscope devices is known to cause numerous variations in contrast which may result from the use of unpolarised and cross-polarised light. Variations in illumination, and noise (\cite{ganster2001melanoma, barata2013dermoscopy}) may also be present, amongst other types of artifacts. Researchers have become increasingly focused on the effect of artifacts in recent years, however, such studies are limited, and tend to focus mainly on the effects of hair (\cite{koehoorn2015hair, bibiloni2017hair}). Studies which focus on detection of melanoma and other skin lesion pathologies have become a significant field of research, especially following the rise in the use of deep learning (\cite{esteva2017skin, brinker2019outperformed, brinker2019classification, fujisawa2019surpasses, pham2020binary, jinnai2020pigmented, jaworek2021sites}). However, despite the acknowledgement of the presence of artifacts in the associated datasets, solutions are either not explored (\cite{mahbod2019hybrid, unver2019lesion}) or are limited (\cite{Pewton_2022_CVPR}).

The most popular methods used to overcome the effect of artifacts on skin lesion images on deep learning models has been by the proposition that artifacts should be removed, often with the use of inpainting methods. Although these studies (\cite{zhou2008artifact, sultana2014removal}) achieved some improvement in accuracy, it is unclear if this process is a friend (removed and inpainted artifacts) or a foe (removed and inpainted important features). Additionally, this process involved localisation and inpainting of artifacts, which is usually computationally expensive and inaccurate as there is no ground truth to evaluate performance. Due to these reasons, focusing on the occurrence of DCA, we propose to superimpose DCA and train a deep learning network to learn these types of artifacts. The aim of this paper is to answer the following questions in binary classification of melanoma and non-melanoma:

\begin{enumerate}
    \item What is the effect of DCA in skin lesion binary classification?
    \item Will inpainting algorithms improve the accuracy of skin lesion binary classification?
    \item Which method provides the best results: inpainted DCA or superimposed synthetic DCA?
    \item Will the deep learning algorithm learn to ignore DCA like the dermatologists?
\end{enumerate}

The main contributions of this paper are as follows:
\begin{itemize}
    \item Introduction of a new DCA split balanced dataset which we make available to the research community. To assist in answering research questions (1) and (2), we curate a new dataset to allow for fair comparison. To date, there are no publicly available DCA split balanced datasets.
    \item A proposed realistic DCA method and compare its performance with binary DCA (\cite{Pewton_2022_CVPR}). We investigate different DCA augmentation techniques to study the effect of DCA inpainting versus the effect of superimposing DCA as in research question (3).
    \item A quantitative measure is proposed to evaluate the visualised activation maps that are commonly used in deep learning research. We measure the differences between deep learning methods when performing inference on images with DCA, and draw a new perspective in handling DCA which can be used in other artifact-related research.
\end{itemize}

\section{Related Work}
\cite{tschandl2019human} conducted experiments using the ISIC datasets (\cite{gutman2016isic, codella2018skin, codella2018isic, tschandl2018ham, combalia2019bcn20000, rotemberg2021dataset}) to compare the diagnostic accuracy of deep learning algorithms with human readers for all clinically relevant types of benign and malignant pigmented skin lesions. Their findings showed that classifiers often had good performance when tested on data that is similar to the training data but performed worse or failed on out of distribution examples. \cite{nauta2022shortcut} observed that artifacts can lead to shortcut learning. Their work focused on detecting and quantifying shortcut learning in trained classifiers for skin cancer diagnosis using the ISIC datasets. They trained a standard VGG16 skin cancer classifier with data split so that elliptical colour patches were present only in the benign images. They inserted colour patches onto images which did not already have them and used inpainting to automatically remove patches from images to assess the effect on predictions. They found that the classifier would partly base its benign predictions on the presence of the coloured patches, and that artificially inserting coloured patches into malignant images resulted in shortcut learning leading to a significant increase in misdiagnosis.

\cite{zand2021preprocessing} conducted experiments to reduce the severity of several types of artifacts in the ISIC-2017 dataset by cropping lesions into rectangles, which reduced the amount of artifacts present in each image. However, although they report good accuracy results (0.8893), this work did not observe the effect of individual artifacts on classification. \cite{cassidy2022isic} trained a wide range of CNN architectures for binary and multi-class classification using the ISIC datasets and observed through the use of Grad-CAM heatmap visualisation that classifiers would frequently focus on artifacts such as air pockets, hair, immersion fluid, measurement overlays, and physical rulers. Such artifacts were shown to have negative effects on classification performance, most critically in cases where melanoma would be misclassified.

Early attempts to address the impact of DCA were conducted by \cite{sultana2014removal}. They performed simple rectangular cropping to remove DCA and inpainting of hairs. However, this may result in the removal of lesion details where the lesion is not centered within the image, or if the lesion details naturally extend beyond the rectangular crop.

\cite{Pewton_2022_CVPR} observed the effects of DCA on the ISIC image datasets. They found that DCA that occupied a large percentage of the image influenced the classification of melanoma vs non-melanoma with a bias towards the melanoma classification. DCA were annotated and dynamic masking and removal methods were proposed. The methods proposed were evaluated with a variety of deep learning architectures, with results showing that the predictive accuracy was comparable, however the network activations showed a large improvement in focus towards lesion regions when images containing DCA were passed through DCA removal methods.

\cite{ramella2021hair} created a method to detect DCA regions in skin lesion images as a pre-processing step which would be performed prior to detection and removal of hair from the images. This was achieved by using the saliency (\cite{ramella2020saliency, ramella2022colour}) and proximity of the DCA to the image frame. Their method was developed as part of a larger workflow to improve hair detection and subsequent removal.

\section{Method}

This section describes the data curation process used to introduce a DCA split balanced dataset, a proposed realistic DCA method, and the experimental settings to evaluate the hypothesis.

\subsection{DCA Split Balanced Dataset}

\begin{figure}[!h]
	\centering
	\begin{tabular}{cc}
	
	    Clean (Train/Val) & Small DCA (Test) \\
		\includegraphics[width=3.5cm,height=3.5cm]{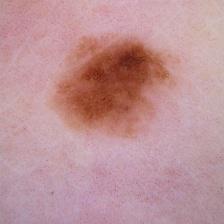} &
		\includegraphics[width=3.5cm,height=3.5cm]{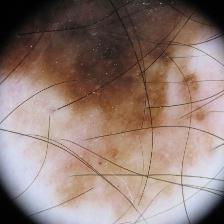} \\
		
		Clean (Train/Val) & Medium DCA (Test) \\
		\includegraphics[width=3.5cm,height=3.5cm]{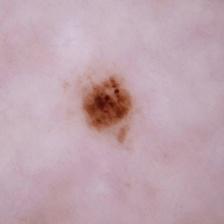} &
		\includegraphics[width=3.5cm,height=3.5cm]{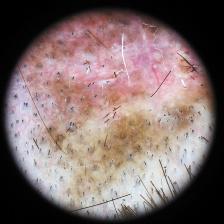} \\
		
		Clean (Train/Val) & Large DCA (Test) \\
		\includegraphics[width=3.5cm,height=3.5cm]{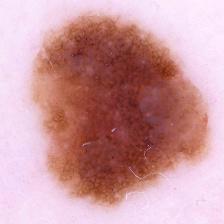} &
		\includegraphics[width=3.5cm,height=3.5cm]{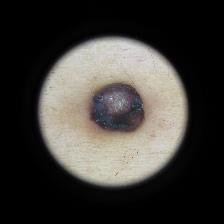} \\
		
		Clean (Train/Val) & Oth DCA (Test) \\
		\includegraphics[width=3.5cm,height=3.5cm]{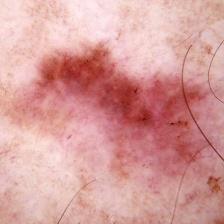} &
		\includegraphics[width=3.5cm,height=3.5cm]{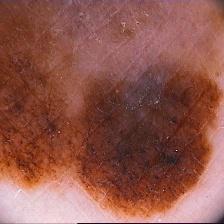} \\

	\end{tabular}
	\caption[]{Example images from the training set (clean) and the testing set (DCA).}
	\label{fig:fig13}
\end{figure}

To further understand the capabilities of deep learning networks for DCA in binary classification (melanoma vs non-melanoma), and the effect of the removal methods, proposed by \cite{Pewton_2022_CVPR}, on classification, a new balanced dataset is formed based on the publicly available dermoscopic datasets (comprising polarised, unpolarised, and a combination of both). Our proposed dataset consists of 10,250 skin lesions images, with a training set without DCA (clean images) and a testing set with DCA. Figure~\ref{fig:fig13} shows examples of images from both the training set (clean images) and the testing set (DCA images). This dataset will enable us to observe the performance of the deep learning algorithm on images with different distributions (\cite{tschandl2019human}). Additionally, the training set without DCA will enable us to superimpose synthetic DCA to study the differences between different DCA types. The testing set with DCA will be used to study the effect of DCA removal and inpainting algorithms (as proposed in \cite{Pewton_2022_CVPR}).


The baseline testing set has been created by including all DCA images from the ISIC balanced dataset proposed by \cite{cassidy2022isic} and separating by DCA size categories (small, medium, large and others) into its own individual testing set. As there are more melanoma images containing DCA than non-melanoma, the non-melanoma category is padded with 1493 images that had been excluded from the original ISIC balanced dataset (\cite{cassidy2022isic}). The images used for padding are manually selected and measured using the DCA masking process proposed by \cite{Pewton_2022_CVPR}. The result of this process produces balanced testing sets where all images in each category contains DCA. 
        
With all DCA images extracted into testing sets, the training and validation sets are unbalanced. These sets contain 1493 fewer melanoma images overall compared to non-melanoma images. In efforts to rebalance the dataset, melanoma images from the Fitzpatrick 17k dataset (\cite{groh2021evaluating}) have been inspected to determine if they are dermatoscopic images and if they contain DCA. Of the 16,529 images contained within the Fitzpatrick 17k dataset, 490 images are annotated as being melanoma by the dataset curator. Of these 490 melanoma images, we annotated 220 of these images to be free from DCA and usable in the DCA Split Balanced Dataset.
      
Following the incorporation of the Fitzpatrick 17k dataset into the training/validation sets, the sets required rebalancing. To rebalance the dataset, all of the melanoma images are shuffled to ensure a good distribution and then split using the holdout method. The training set contains 90\% of the melanoma images, whilst the validation set contains the further 10\%. As per the melanoma images, the non-melanoma images are shuffled. As there are many more non-melanoma images than melanoma - the extra non-melanoma images are removed to leave an equal number of images per class. The non-melanoma images are then split in the same way as the melanoma images.

Table~\ref{tab:tab01} shows the final distribution of balanced training, validation and testing sets, with 5512, 614 and 4124 images, respectively. Although more images within the training sets would be desirable, there are a limited number of publicly available melanoma datasets containing DCA-free images.

\begin{table}[htbp]
    \centering
    \caption{Summary of the DCA Split Balanced Dataset which contains a total of 10,250 images. Mel - melanoma; Non-Mel - Non-melanoma.}
    \scalebox{1.00}{
    \begin{tabular}{lllllll}
        \cline{2-7}
            \multicolumn{1}{c}{} & \multicolumn{2}{c}{\textbf{Training Set}} & \multicolumn{4}{c}{\textbf{Testing Set (DCA Sizes)}} \\
        \cline{2-7}
            \multicolumn{1}{c}{} & \textbf{\textit{Train}} & \textbf{\textit{Val}} & \textbf{\textit{Small}} & \textbf{\textit{Medium}} & \textbf{\textit{Large}} & \textbf{\textit{Other}} \\
        
        \hline
            \textbf{Mel} & 2756 & 307 & 909 & 488 & 423 & 242 \\ 
        \hline
            \textbf{Non-Mel} & 2756 & 307 & 909 & 488 & 423 & 242 \\
        \hline
    
    \end{tabular}
    }
    \label{tab:tab01}
\end{table}

\subsection{Recreating a Realistic DCA}

\begin{figure*}[!htbp]
	\centering
	\begin{tabular}{c}
		\includegraphics[scale=0.32]{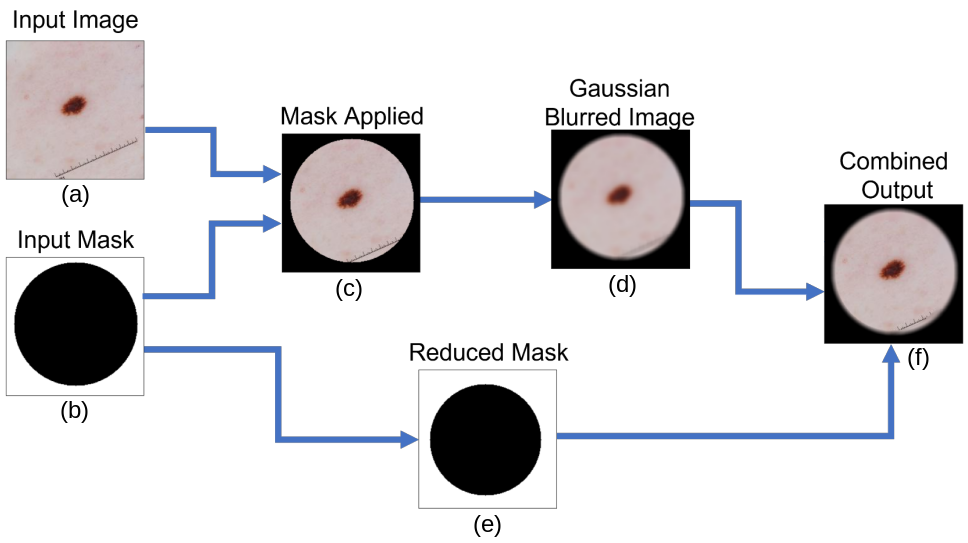}
	\end{tabular}
	\caption[test]{Illustration of our proposed realistic DCA creation process. First, the input image (a) and input mask (b) are combined to generate a binary mask image (c). Then, the binary mask is passed through a Gaussian blur (d). Finally, the input image and the Gaussian blur image are combined using a reduced mask (e) as a filter. Any pixels in the filter that is contained within the circle is replaced with the original input image, and pixels outside are replaced with the Gaussian blurred image equivalent. The final output is the combined output (f).}
	\label{fig:fig06}
\end{figure*}

Our experiments utilise two types of synthetic DCA: (1) binary DCA as proposed by \cite{Pewton_2022_CVPR}, and (2) a proposed more realistic DCA. Figure~\ref{fig:fig06} illustrates the processes to create a realistic DCA.

        
Building on the DCA masks extraction process by \cite{Pewton_2022_CVPR}, the mask is applied to an image which is then processed using a Gaussian blur. Once a blurred image is generated, the original centre point and radius is extracted from the original DCA mask. The radius of the circle is reduced to determine the gradient of the transitional area between the image and DCA. Using this modified radius, a new mask is generated. The newly generated mask is used to determine the area of a new image which should contain the data from the blurred image, and the remaining area to contain data from the original unblurred image. The final stage involves the merging of the Gaussian blurred image and the reduced mask.

\subsection{Classification}
To evaluate the effect of DCA on binary classification, InceptionResNetV2 (\cite{szegedy2017inception}) was selected for our experiments, as this network was the best overall performing network in our prior work on DCA (\cite{Pewton_2022_CVPR}). Our intention is not to produce the best classification algorithm, but to investigate on the new strategy of our proposed dataset and training approaches. The three models are trained by using three types of training and validation sets: (1) clean (original images without DCA), (2) binary DCA (original images superimposed with binary DCA), and (3) realistic DCA (original images superimposed with realistic DCA).
    
The models were trained with no pre-trained weights to a maximum of 200 epochs, and early stopping after 10 epochs if no validation accuracy increase is achieved. The models were trained using a batch size of 64 with stochastic gradient descent as the optimiser. The model exhibiting the highest validation accuracy was saved. No model fine tuning was completed to ensure fairness and equality between the three models trained. The hardware configuration used to train all of the networks was an AMD Ryzen 7 3700X 8-core 16-thread 4.4GHz CPU with 16GB DDR4 3000MHz Dual-Channel RAM and an NVIDIA Geforce RTX 3090 FE 24GB GDDR6X GPU. The software configuration used was Python 3.9.7, TensorFlow GPU 2.9.0-dev20220203, CUDA 11.2.1, and cuDNN 8.1 running on Windows 10.

Table~\ref{tab:tab04} shows the overall model accuracy achieved with the validation set across each of the models trained.
            
    \begin{table}[!htbp]
        \centering
        \caption{Trained model metrics for each training set. Acc - accuracy, AUC - Area Under the Curve.}
        \scalebox{1.00}{
        \begin{tabular}{llll}
            \hline
            \textbf{Training Dataset} & \textbf{Best Epoch} & \textbf{Val Acc} & \textbf{Val AUC} \\
            \hline
            \hline
            Clean            & 29         & 0.82    & 0.88\\
            \hline
            Binary DCA       & 15         & 0.81    & 0.88 \\
            \hline
            Realistic DCA    & 30         & 0.81    & 0.89\\
            \hline
        \end{tabular}
        }
        \label{tab:tab04}
    \end{table}

\subsection{Experiments}
To provide an in depth understanding of the effect of DCA, we conduct three experiments. In Experiment I, we test the clean model on the testing set with DCA, the testing set with DCA inpainted by Navier-Stokes (NS), and the testing set with DCA inpainted by Telea. Experiment II studies the effect of superimposed synthetic DCA (binary DCA and realistic DCA) training models on the testing set, i.e., the binary DCA model and realistic DCA model are tested on the testing set. Due to the varied size of DCA, we report the detailed results according to each category. Experiment III investigates the performance of Binary and Realistic DCA models on the testing set with DCA removal and inpainted by NS and Telea.

\subsection{Performance Metrics}
For evaluation on the binary classification task, common performance metrics including Accuracy (Acc), True Positive Rate (TPR, also known as sensitivity), True Negative Rate (TNR, also known as specificity), Precision, F1-Score, and Area under the Receiver Operating Curve (AUC) are used. To further elaborate on the differences between the network performance of the different models, Grad-CAM (\cite{selvaraju2019gradcam}) is used to extract the network activation gradients from the last convolutional layer of each network. The gradients extracted produce a heatmap to show which parts of an image the network focuses on to determine the classification result. For our experiments, a class implementation of Grad-CAM created by \cite{rosebrock_2020} has been used. Within the resulting heatmaps, the bright yellow regions are areas used heavily by the network for prediction.

We propose a new quantitative method by introducing intensity measures for prediction activation heatmaps. To quantify the area targeted by the heatmaps produced for the Grad-CAM experiments, the heatmaps for all test sets across all networks are extracted from the test predictions. The corresponding mask for each of the images is used to segregate the two areas in the image - the external section of the image is the DCA region, and the internal section of the image is the area in which the lesion resides.

As the areas of the image that are focused on by the network are brighter than those that are not - the brightness values make it possible to measure the difference that each method has made on the corresponding heatmaps. Using the internal and external areas of the heatmap image, the root mean square (RMS) contrast and the average brightness value is calculated for each heatmap. This process was completed using the ImageStat method which is part of the Pillow Python library (\cite{lund2020pillow}). RMS contrast is not dependant on the spatial distribution or the angular frequency content of contrast in the image, and is defined as the standard deviation of pixel intensities (\cite{peli1990contrast}). The relevant mathematical expression for RMS contrast is as follows:

\begin{equation}
RMS = \sqrt{\frac{1}{MN}\sum_{i=0}^{N-1}\sum_{j=0}^{M-1}(I_{ij}-\bar{I})^{2}}
\end{equation}
where intensities $I_{ij}$ are the $i$-th $j$-th element of the two-dimensional image of size $M$ by $N$. $\bar{I}$ is the average intensity of all pixel values in the image. The image $I$ is assumed to have pixel intensities normalized in the range of [0, 1].


\section{Results}
This section presents the results of our proposed synthetic DCA and experiment I-III.

\subsection{Synthetic DCA}
\begin{figure}[!htbp]
        	\centering
        	\begin{tabular}{ccc}
        	
        	      Small & Medium & Large \\
        	      
        	      \rotatebox{90}{Original}
        	      \includegraphics[width=2.5cm,height=2.5cm]{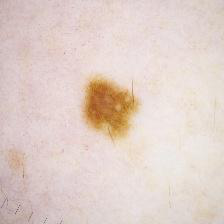} & 
        	      \includegraphics[width=2.5cm,height=2.5cm]{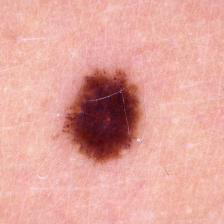} & 
        	      \includegraphics[width=2.5cm,height=2.5cm]{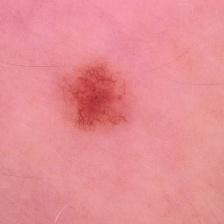} \\
        	      
        	      \rotatebox{90}{Binary DCA}
        	      \includegraphics[width=2.5cm,height=2.5cm]{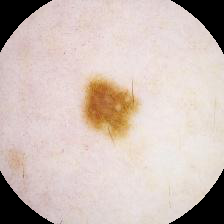} & 
        	      \includegraphics[width=2.5cm,height=2.5cm]{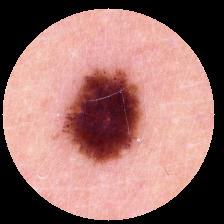} & 
        	      \includegraphics[width=2.5cm,height=2.5cm]{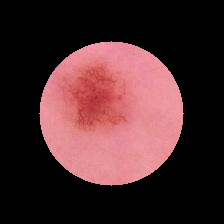} \\
        	      
        	      \rotatebox{90}{Realistic DCA}
        	      \includegraphics[width=2.5cm,height=2.5cm]{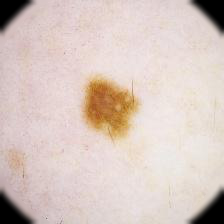} & 
        	      \includegraphics[width=2.5cm,height=2.5cm]{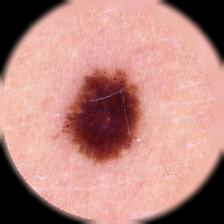} & 
        	      \includegraphics[width=2.5cm,height=2.5cm]{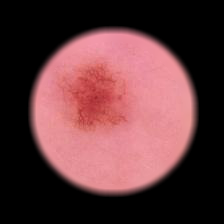} \\
        	\end{tabular}
        	\caption[]{Illustration of different superimposed DCA sizes on skin lesion images. The first row shows the original images from the training set. The second row shows the superimposed binary DCA, and the third row shows the superimposed realistic DCA.}
        	\label{fig:fig07}
        \end{figure}

Figure~\ref{fig:fig07} shows an example of superimposed the binary DCA and realistic DCA on a clean image from training set.

\begin{figure*}[!htbp]
        	\centering
        	\begin{tabular}{ccc}
        	
        		\includegraphics[width=3.6cm,height=3.6cm]{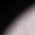} &
        		\includegraphics[width=3.6cm,height=3.6cm]{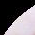} &
        		\includegraphics[width=3.6cm,height=3.6cm]{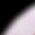} \\
        		True DCA & Binary DCA & Realistic DCA \\
        
        	\end{tabular}
        	\caption[]{A visual comparison (close-up 35$\times$35 pixels) of true DCA and sythnetic DCAs. (Left) True DCA from the test set; (Middle) Binary DCA superimposed onto an image in the training set as proposed by \cite{Pewton_2022_CVPR}; (Right) Our proposed realistic DCA on an image in the training set. Note that visually, our proposed realistic DCA closely resembles the true DCA.}
        	\label{fig:fig12}
        \end{figure*}

As can be seen in Figure~\ref{fig:fig07}, the proposed realistic DCA method produces visually effective results for each of the DCA sizes. Figure~\ref{fig:fig12} presents a close up visual patch comparison (35x35 px) of a true DCA (real DCA from the testing set), a binary DCA superimposed on an image from training set, and a realistic DCA superimposed on a similar image from training set.
It can be clearly seen that the realistic DCA has a smooth transition into the DCA region from the image, much like the true DCA image whereas the binary DCA forms a distinct solid boundary between lesion and DCA region.

\subsection{Experiment I: The effect of DCA on skin lesions classification}
Table~\ref{tab:tab04} shows the evaluation metrics for the clean model (without DCA) on the testing set (with DCA). Due to the composition of the training set, it is expected that the results will be mostly predicted as melanoma, with high sensitivity (TPR) but low specificity (TNR), as illustrated in the first row of Table~\ref{tab:tab04}. This result is aligned with previous research which indicates that superficial spreading melanoma are dark brown or black in appearance (\cite{rose1998lesions}). After removal of DCA by inpainted with Navier-Stokes and Telea, we observed marginal improvement to TNR, Precision, F1-Score and AUC.
        
        \begin{table}[!htbp]
            \centering
            \caption{The performance of the clean model on the test set (Original); test set with DCA inpainted by Navier-Stokes (NS); and test set with DCA inpainted by Telea (Telea).}
            \scalebox{1.00}{
            \begin{tabular}{lllllll}
                \hline
                 \textbf{Test Set}      & \multicolumn{6}{l}{\textbf{Metrics}} \\
                \cline{2-7}
                                    & \textbf{Acc}  & \textbf{TPR}  & \textbf{TNR}  & \textbf{Precision} & \textbf{F1} & \textbf{AUC} \\
                \hline
                 Original          & 0.57 & \textbf{0.90} & 0.23 & 0.54& \textbf{0.68} & 0.61   \\
                 NS                & \textbf{0.59} & 0.86 & 0.31 & \textbf{0.56}& \textbf{0.68} & 0.64   \\
                 Telea             & \textbf{0.59} & 0.84 & \textbf{0.34} & \textbf{0.56}& 0.67 & \textbf{0.65}   \\
                \hline 

                \hline
            \end{tabular}
            }
            \label{tab:tab04}
        \end{table}

To further support the efficiency of DCA removal methods on the testing set, the performance of all test sets with different DCA sizes are compared. Table~\ref{tab:tab02} shows the performance of the clean model evaluated on the test set.
        \begin{table}[!htbp]
            \centering
            \caption{The effect of DCA according to DCA size on the performance of the clean model on the test set (Original); test set with DCA inpainted by Navier-Stokes (NS); and test set with DCA inpainted by Telea (Telea).}
            \scalebox{1.00}{
            \begin{tabular}{lllllll}
                \hline
                \textbf{Test Set - DCA Size} & \multicolumn{6}{l}{\textbf{Metrics}} \\
                \cline{2-7}
                  & \textbf{Acc} & \textbf{TPR} & \textbf{TNR} & \textbf{Precision} & \textbf{F1} & \textbf{AUC} \\
                \hline
                \hline
                 Original - small & 0.59 & 0.86 & 0.32 & 0.56 & 0.68 & 0.63 \\
                  NS - small & 0.58 & 0.87 & 0.30 & 0.55 & 0.67 & 0.62 \\
                  Telea - small & 0.58 & 0.86 & 0.30 & 0.55 & 0.67 & 0.62 \\
                \cline{1-7}
                  Original - medium & 0.57 & \textbf{0.91} & 0.24 & 0.54 & 0.68 & 0.64 \\
                  NS - medium & 0.58 & 0.90 & 0.26  & 0.55 & 0.68 & 0.65\\
                  Telea - medium & \textbf{0.59} & 0.88 & \textbf{0.30}  & \textbf{0.56} & 0.68 & \textbf{0.66}\\
                \cline{1-7}
                  Original - large & 0.51 & \textbf{0.99} & 0.01& 0.50 & 0.67 & 0.58 \\
                  NS - large & \textbf{0.62} & 0.81 & 0.43 & \textbf{0.59}& \textbf{0.68} & 0.67 \\
                  Telea - large & 0.61 & 0.72 & \textbf{0.50} & \textbf{0.59}& 0.65 & \textbf{0.68} \\
                \cline{1-7}
                  Original - other & 0.58 & 0.90 & 0.26  & 0.55 & 0.67 & 0.65\\
                  NS - other & 0.57 & 0.87 & 0.27& 0.54 & 0.67 & 0.65 \\
                  Telea - other & 0.57 & 0.87 & 0.27& 0.54 & 0.67 & 0.64 \\
                \cline{1-7}

            \end{tabular}
            }
            \label{tab:tab02}
        \end{table}
The results are comparable across the `small' and `other' DCA sizes, however the medium and large DCA sizes show a more notable increase in accuracy, TNR and precision. The largest accuracy increase is seen in the large DCA test sets where the NS inpainting method achieves 11\% greater accuracy compared to the baseline test set containing original DCA images. Another observation is that the Original large DCA has 0.01 TNR, which means almost all the large DCA were classified as melanoma. With Telea inpainting, the TNR improved 49\%, with 0.50 TNR.

\subsection{Experiment II: The effect of superimposed DCA}
When comparing the overall accuracy performance of the original test set across the 3 networks in Table~\ref{tab:tab04_1}, an accuracy increase of 3\% is seen when the images are predicted using the model trained with binary DCA, and an increase of 4\% when predicted with the model trained with realistic DCA. The performances across the original test set also shows a significant increase in TNR on the realistic DCA model - meaning that there is less of a blanket classification of melanoma across the network. Due to the variance of predictions made, this model also suffers from a reduction in TPR when compared to the predictions made with the cleanly trained model. 

        \begin{table}[!htbp]
            \centering
            \caption{The performance of all trained networks (clean model, superimposed binary DCA model and superimposed realistic DCA model) on the test set.}
            \scalebox{1.00}{
            \begin{tabular}{lllllll}
                \hline
                \textbf{Model}                & \multicolumn{6}{l}{\textbf{Metrics}} \\
                \cline{2-7}
                                    & \textbf{Acc}  & \textbf{TPR}  & \textbf{TNR} & \textbf{Precision} & \textbf{F1}   & \textbf{AUC}  \\
                \hline
                \hline
                Clean                   & 0.57 & 0.90 & 0.23 & 0.54 & 0.68 & 0.61  \\
                \hline
                Binary DCA             & 0.60 & \textbf{0.91} & 0.29 & 0.56 & \textbf{0.70} & \textbf{0.66}  \\
                \hline
                Realistic DCA         & \textbf{0.61} & 0.73 & \textbf{0.49} & \textbf{0.59} & 0.65 & \textbf{0.66}  \\
                \hline 

            \end{tabular}
            }
            \label{tab:tab04_1}
        \end{table}

\begin{figure}[!htbp]
            \centering
            \includegraphics[scale=0.283]{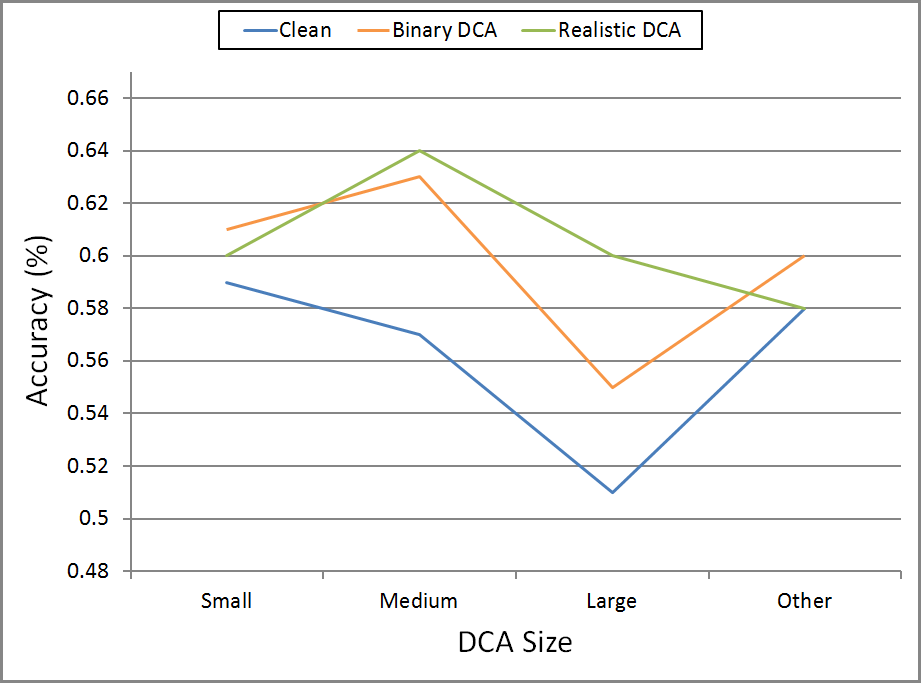}
            \caption{Model accuracy for small, medium, large, and other DCA sizes.}
            \label{fig:fig01}
        \end{figure}

To further investigate the performance differences across the different models and test sets, metrics were then generated for all test sets across all DCA sizes. Table~\ref{tab:tab03} shows the metrics that were generated for the original DCA sets across each of the models and Figure~\ref{fig:fig01} shows a line graph comparing the model accuracy performance between the different DCA sizes. Full model performance metrics generated across the three networks are available on our GitHub repository (link will be provided following acceptance of paper).
    
        \begin{table}[!htbp]
            \centering
            \caption{The performance of all trained networks (clean model, superimposed binary DCA model and superimposed realistic DCA model) on the test set with different DCA sizes.}
            \scalebox{1.00}{
            \begin{tabular}{llllllll}
                \hline
                \textbf{DCA Size}      & \textbf{Model}         & \multicolumn{6}{l}{\textbf{Metrics}} \\
                \cline{3-8}
                              &               & \textbf{Acc}  & \textbf{TPR}  & \textbf{TNR}  & \textbf{Precision} & \textbf{F1}   & \textbf{AUC}  \\
                \hline
                \hline
                Small  & Clean         & 0.59 & 0.86 & 0.32 & 0.56& 0.68 & 0.63 \\
                              & Binary DCA    & \textbf{0.61} & \textbf{0.90} & 0.33 & 0.57& \textbf{0.70} & \textbf{0.67} \\
                              & Realistic DCA & 0.60 & 0.85 & \textbf{0.35}& \textbf{0.57} & 0.68 & 0.65 \\
                \hline
                Medium& Clean   & 0.57 & 0.91 & 0.24 & 0.54& 0.68 & 0.64 \\
                              & Binary DCA    & 0.63 & \textbf{0.94} & 0.31 & 0.58& \textbf{0.72} & 0.68 \\
                              & Realistic DCA & \textbf{0.64} & 0.75 & \textbf{0.53} & \textbf{0.62}& 0.68 & \textbf{0.70} \\
                \hline
                Large  & Clean         & 0.51 & \textbf{0.99} & 0.01 & 0.50& 0.67 & 0.58 \\
                              & Binary DCA    & 0.55 & 0.96 & 0.13& 0.53 & \textbf{0.68} & 0.62 \\
                              & Realistic DCA & \textbf{0.60} & 0.39 & \textbf{0.80}& \textbf{0.66} & 0.50 & \textbf{0.63} \\
                \hline
                Other  & Clean         & 0.58 & \textbf{0.90} & 0.26  & 0.55 & \textbf{0.67} & 0.65\\
                              & Binary DCA    & \textbf{0.60} & 0.83 & \textbf{0.36}  & \textbf{0.57}& \textbf{0.67} & \textbf{0.67}\\
                              & Realistic DCA & 0.58 & 0.81 & 0.35  & 0.56& 0.66 &         
    0.65\\
                \hline 

                \hline
            \end{tabular}
            }
            \label{tab:tab03}
        \end{table}
        
As can be seen in Table~\ref{tab:tab03} and Figure~\ref{fig:fig01}, the overall accuracy has increased for each of the test sets when the network is trained using both binary and realistic DCA. The largest accuracy increase can be seen for the large DCA test sets where the network trained with realistic DCA shows an increase in accuracy of 9\%. Significant increases can also be seen for TNR on medium and large augmented DCA models indicating that more non-melanoma images were correctly classified.

Figure \ref{fig:fig01} shows that network accuracy performance was equal to or improved from the baseline DCA set tested on the cleanly trained network. The binary DCA model shows the best overall accuracy for both the small and `other' sized DCA test sets, the realistic DCA showed the best overall accuracy for the medium sized test set, and the Telea inpainted images on the clean model showed the best overall accuracy for the large test set (Realistic DCA). Although both of the results for the large testing set from the models trained on synthetic DCA images improves from the baseline performance - the models still appear to be hindered due to the large DCA size.

\begin{figure}[!h]
	\centering
	\begin{tabular}{ccc}
	
	      & Original Image &  \\
	      
	      & \includegraphics[width=2.5cm,height=2.5cm]{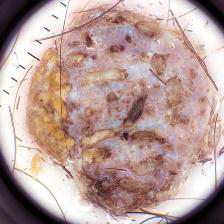} & \\
	    
	      Clean & Binary DCA & Realistic DCA \\
	      
		 \includegraphics[width=2.5cm,height=2.5cm]{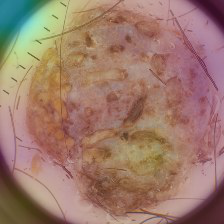} & 
		 \includegraphics[width=2.5cm,height=2.5cm]{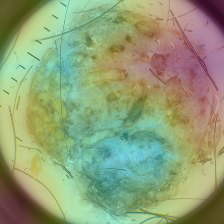} & 
		 \includegraphics[width=2.5cm,height=2.5cm]{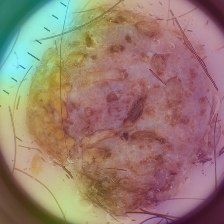} \\
		 
		 \includegraphics[width=2.5cm,height=2.5cm]{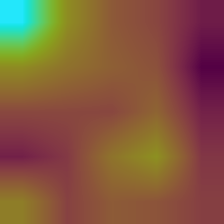} & 
		 \includegraphics[width=2.5cm,height=2.5cm]{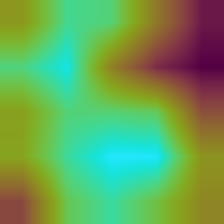} & 
		 \includegraphics[width=2.5cm,height=2.5cm]{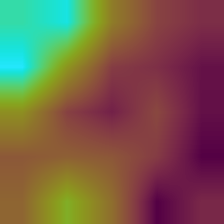} \\

	\end{tabular}
	\caption[]{Illustration of Grad-CAM results on a small DCA image for clean, binary, and realistic DCA models.}
	\label{fig:fig02}
\end{figure}

Figure~\ref{fig:fig02} shows the output results of a small DCA image across each of the three networks. The activations on the heatmap generated from the clean model focus largely on the area of the lesion, however, the top-left corner also exhibits significant focus. When comparing the results from the cleanly trained model with the results generated from the binary and realistic DCA trained models, the corner region becomes less of an area of interest when DCA is used for training. Although improvement can be seen in the activation area where the lesion is located, activations are still partially present on the dark regions. The network showing the best activations for small DCA images is the realistic DCA model.

\begin{figure}[!htbp]
	\centering
	\begin{tabular}{ccc}
	
	      & Original Image &  \\
	      
	      & \includegraphics[width=2.5cm,height=2.5cm]{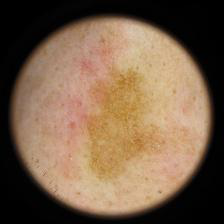} & \\
	    
	      Clean & Binary DCA & Realistic DCA \\
	      
		 \includegraphics[width=2.5cm,height=2.5cm]{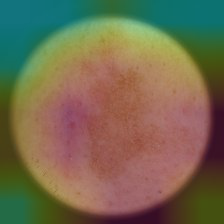} & 
		 \includegraphics[width=2.5cm,height=2.5cm]{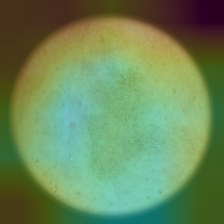} & 
		 \includegraphics[width=2.5cm,height=2.5cm]{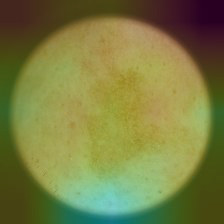} \\
		 
		 \includegraphics[width=2.5cm,height=2.5cm]{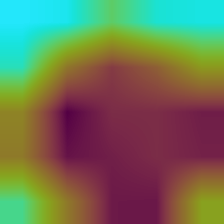} & 
		 \includegraphics[width=2.5cm,height=2.5cm]{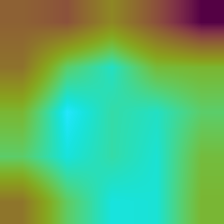} & 
		 \includegraphics[width=2.5cm,height=2.5cm]{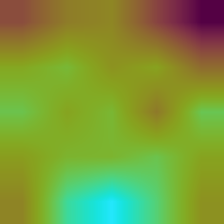} \\

	\end{tabular}
	\caption[]{Illustration of Grad-CAM results on a medium DCA image for clean, binary, and realistic DCA models.}
	\label{fig:fig03}
\end{figure}

Figure~\ref{fig:fig03} shows the output results of a medium DCA image across each of the three networks. The network activations on the cleanly trained model mostly focus on the upper corners of the image for medium sized DCA. When compared to the results from the binary and realistic DCA models, we observe that the network is able to effectively ignore the majority of the DCA region. The activations for the binary and realistic DCA models are similar, though the binary DCA activations appear to encapsulate the lesion more finely and ignore more of the DCA region than the model trained on realistic DCA. Both the binary and realistic DCA model results show a large improvement in the focus of activations for medium DCA images.
        


\begin{figure}[!htbp]
	\centering
	\begin{tabular}{ccc}
	
	      & Original Image &  \\
	      
	      & \includegraphics[width=2.5cm,height=2.5cm]{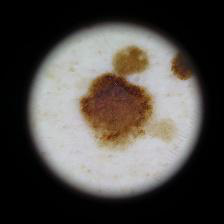} & \\
	    
	      Clean & Binary DCA & Realistic DCA \\
	      
		 \includegraphics[width=2.5cm,height=2.5cm]{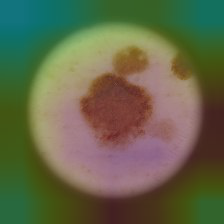} & 
		 \includegraphics[width=2.5cm,height=2.5cm]{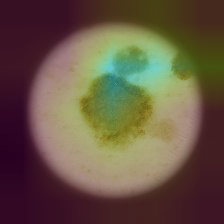} & 
		 \includegraphics[width=2.5cm,height=2.5cm]{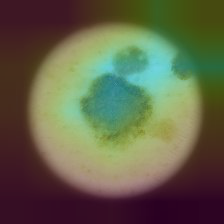} \\
		 
		 \includegraphics[width=2.5cm,height=2.5cm]{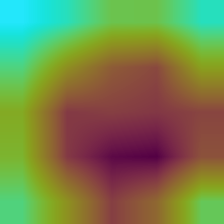} & 
		 \includegraphics[width=2.5cm,height=2.5cm]{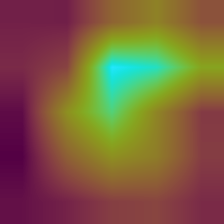} & 
		 \includegraphics[width=2.5cm,height=2.5cm]{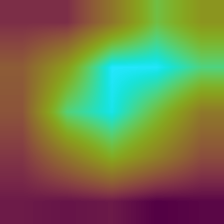} \\

	\end{tabular}
	\caption[]{Illustration of Grad-CAM results on a large DCA image for clean, binary, and realistic DCA models.}
	\label{fig:fig04}
\end{figure}
        
Figure~\ref{fig:fig04} shows similar results to Figure~\ref{fig:fig03} for large DCA. The model trained on a clean dataset almost entirely focuses on the DCA region, while the models trained on binary DCA and realistic DCA show a large improvement on the focus towards the activations. The results for both DCA trained models are both similar, with the binary DCA image showing a smaller surface area of activations as opposed to the realistic DCA model. Both DCA models show significant improvements on large DCA images when compared to the results from previous experiments using small and medium DCA.

\begin{figure}[!h]
	\centering
	\begin{tabular}{ccc}
	
	      & Original Image &  \\
	      
	      & \includegraphics[width=2.5cm,height=2.5cm]{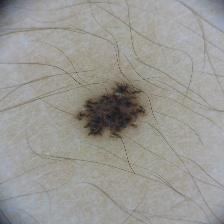} & \\
	    
	      Clean & Binary DCA & Realistic DCA \\
	      
		 \includegraphics[width=2.5cm,height=2.5cm]{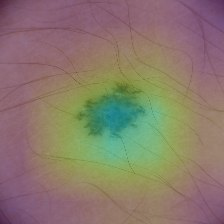} & 
		 \includegraphics[width=2.5cm,height=2.5cm]{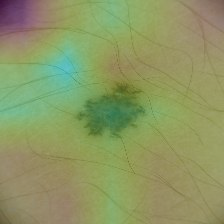} & 
		 \includegraphics[width=2.5cm,height=2.5cm]{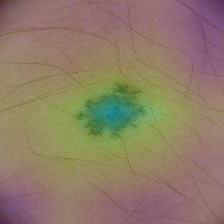} \\
		 
		 \includegraphics[width=2.5cm,height=2.5cm]{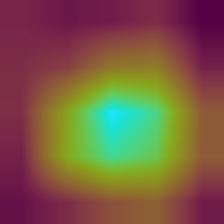} & 
		 \includegraphics[width=2.5cm,height=2.5cm]{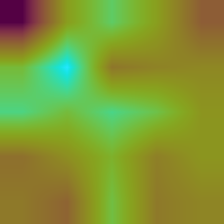} & 
		 \includegraphics[width=2.5cm,height=2.5cm]{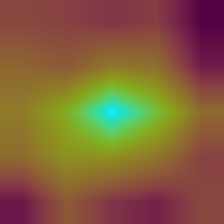} \\

	\end{tabular}
	\caption[]{Illustration of Grad-CAM results on a DCA image from the `other' category for clean, binary, and realistic DCA models.}
	\label{fig:fig05}
\end{figure}
        
Figure~\ref{fig:fig05} shows the output results for a `other' DCA image across each of the three networks. Strong activations for both the model trained on the clean dataset and the model trained on realistic DCA are clearly visible. The heatmap generated from the model trained on binary DCA images completely loses focus on the lesion area. Between the results from the clean model and the realistic DCA model, the activations in the clean model are more strongly focused on the area of interest.
        

\subsection{Experiment III: DCA removal testing set vs Superimposed Synthetic DCA training set}

To understand the differences in model performance between the DCA removal processes and superimposed synthetic DCA in the training process, it is necessary to cross compare the results produced across each of the models. From Experiment I and II, we observed that both approaches resulted in improved accuracy, TNR and precision. Therefore, we compare their performance on different DCA sizes based on the best TNR from each approach. Table~\ref{tab:bestTNR} compares the results. Overall, the superimposed synthetic DCA training model performed the best in accuracy, TNR, and precision. However, the method using inpainted achieved a superior result in TPR. 

       \begin{table}[!htbp]
            \centering
            \caption{The performance of the best TNR from the inpainted method and the superimposed method on the test set with different DCA sizes.}
            \scalebox{1.00}{
            \begin{tabular}{llllllll}
                \hline
                \textbf{DCA Size}      & \textbf{Method}         & \multicolumn{6}{l}{\textbf{Metrics}} \\
                \cline{3-8}
                              &               & \textbf{Acc}  & \textbf{TPR}  & \textbf{TNR}  & \textbf{Precision} & \textbf{F1}   & \textbf{AUC}  \\
                \hline
                \hline
                Small  
                            & Inpainted  & 0.58 & \textbf{0.87} & 0.30 & 0.55 & 0.67 & 0.62 \\
                              & Superimposed & \textbf{0.60} & 0.85 & \textbf{0.35}& \textbf{0.57} & \textbf{0.68} & \textbf{0.65} \\
                \hline
                Medium 
                              & Inpainted     & 0.59 & \textbf{0.88} & 0.30  & 0.56 & 0.68 & 0.66\\
                              & Superimposed & \textbf{0.64} & 0.75 & \textbf{0.53} & \textbf{0.62}& 0.68 & \textbf{0.70} \\
                \hline
                Large  
                              & Inpainted    & \textbf{0.61} & \textbf{0.72} & 0.50 & 0.59& \textbf{0.65} & \textbf{0.68} \\
                              & Superimposed & 0.60 & 0.39 & \textbf{0.80}& \textbf{0.66} & 0.50 & 0.63 \\
                \hline
                Other  
                              & Inpainted    & 0.57 & \textbf{0.87} & 0.27& 0.54 & 0.67 & 0.65 \\
                              & Superimposed& \textbf{0.60} & 0.83 & \textbf{0.36}  & \textbf{0.57}& 0.67 & \textbf{0.67}\\
                \hline 

                \hline
            \end{tabular}
            }
            \label{tab:bestTNR}
        \end{table}
        
\begin{figure*}[!htbp]
	\centering
	\begin{tabular}{c}
		\includegraphics[scale=0.30]{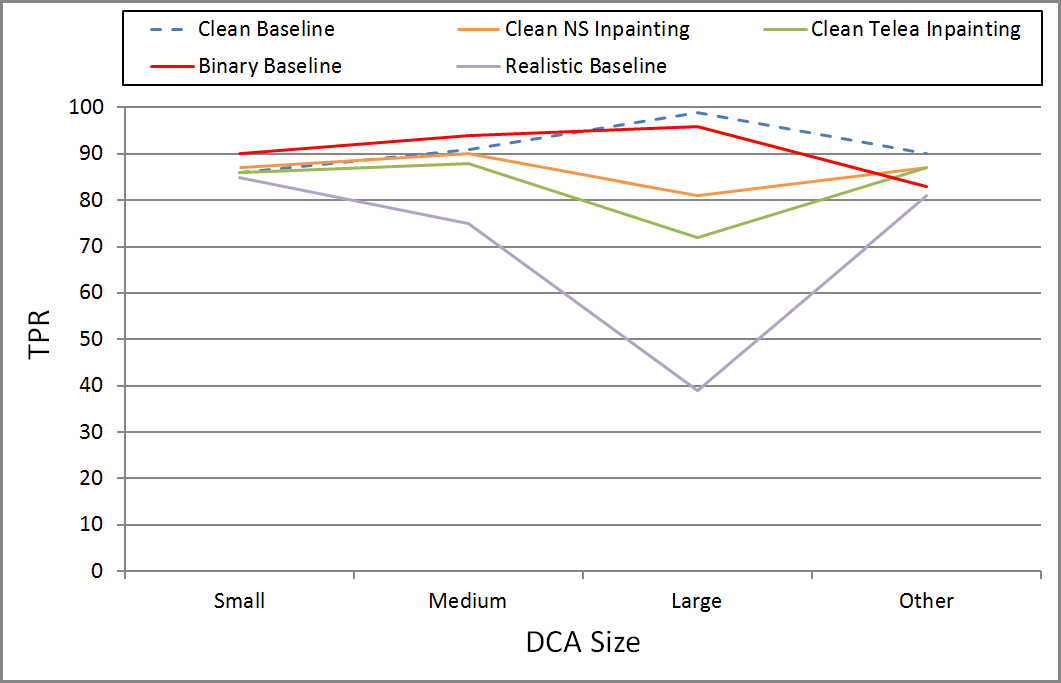}
	\end{tabular}
	\caption[test]{TPR plots for result metrics generated across each trained network.}
	\label{fig:tpr}
\end{figure*}

\begin{figure*}[!htbp]
	\centering
	\begin{tabular}{c}
		\includegraphics[scale=0.30]{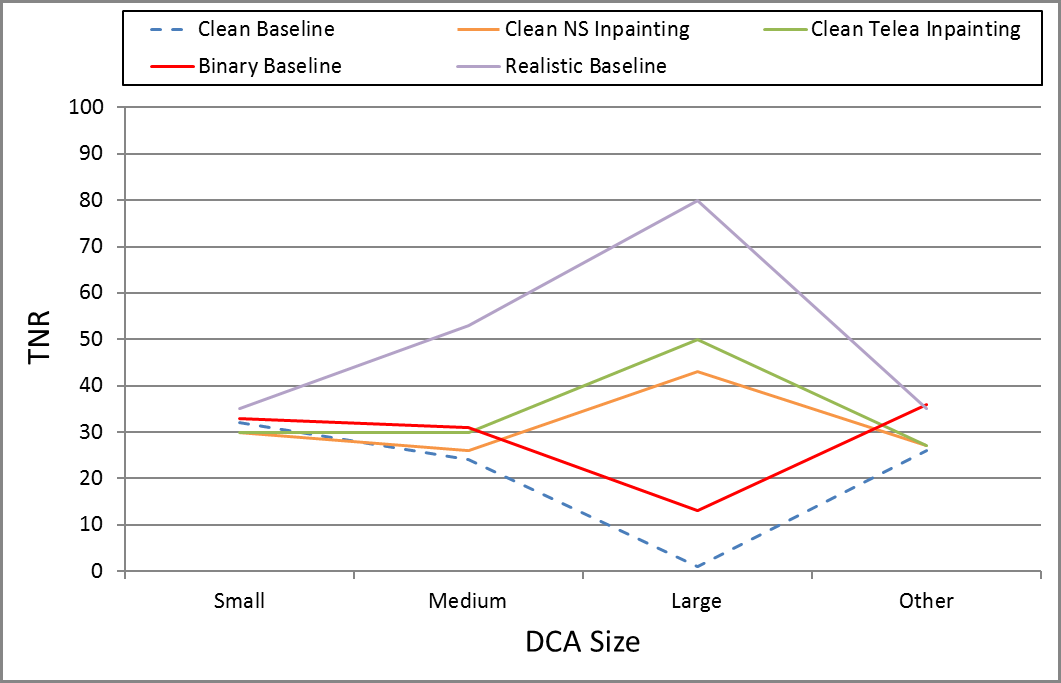}
	\end{tabular}
	\caption[test]{TNR plots for result metrics generated across each trained network.}
	\label{fig:tnr}
\end{figure*}

Figure~\ref{fig:tpr} and Figure~\ref{fig:tnr} show the different results generated for the inpainted DCA images using the Navier-Stokes and Telea inpainting methods against the results generated for DCA images tested within the models trained on superimposed synthetic DCA images. 

It can be seen that the results generated for the inpainted test sets are comparable to the results generated on the DCA test sets on models trained with synthetic DCA. Both methods produce similar accuracy improvements overall. In terms of TPR, the binary baseline model outperforms the other models for small and medium DCA sizes, while the clean baseline model performs best for large and other models. The realistic DCA model shows the largest decrease in TPR, however it also shows the largest increase in TNR. This suggests that the model is able to identify more features within non-melanoma images. The clean and binary models show the most notable drops in TNR for large DCA, both with $<0.20$ TNR, a difference of $>0.60$ compared to the best performing model (realistic DCA).

\begin{figure*}[!htbp]
        	\centering
        	\begin{tabular}{ccccc}
        	
        	     CM & NS & Tel & Bin & Real \\
        	    
        	    \rotatebox{90}{S-DCA \textcolor{white}{Lg}}
        		 \includegraphics[width=1.8cm,height=1.8cm]{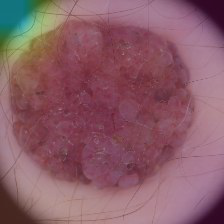} & 
        		 \includegraphics[width=1.8cm,height=1.8cm]{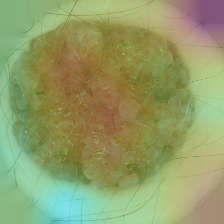} & 
        		 \includegraphics[width=1.8cm,height=1.8cm]{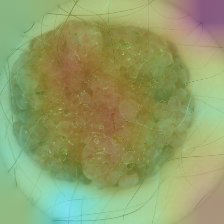} & 
        		 \includegraphics[width=1.8cm,height=1.8cm]{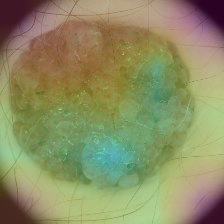} & 
        		 \includegraphics[width=1.8cm,height=1.8cm]{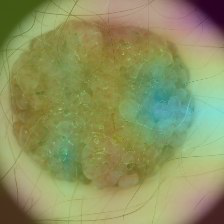} \\
        		
        		\rotatebox{90}{\textcolor{white}{Lg}}
        		 \includegraphics[width=1.8cm,height=1.8cm]{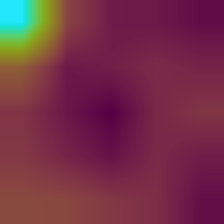} & 
        		 \includegraphics[width=1.8cm,height=1.8cm]{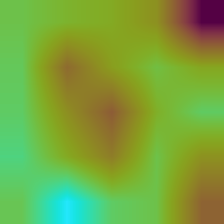} & 
        		 \includegraphics[width=1.8cm,height=1.8cm]{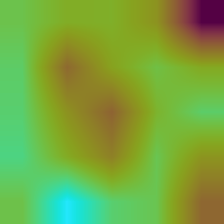} & 
        		 \includegraphics[width=1.8cm,height=1.8cm]{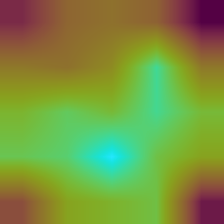} & 
        		 \includegraphics[width=1.8cm,height=1.8cm]{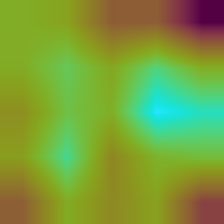} \\
        		
        		\rotatebox{90}{M-DCA \textcolor{white}{Lg}}
        		 \includegraphics[width=1.8cm,height=1.8cm]{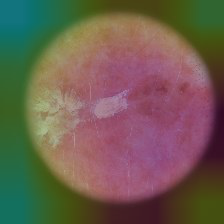} & 
        		 \includegraphics[width=1.8cm,height=1.8cm]{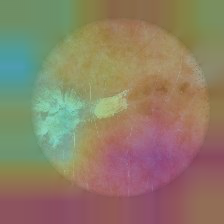} & 
        		 \includegraphics[width=1.8cm,height=1.8cm]{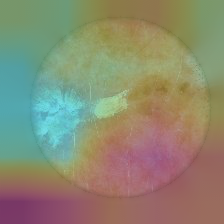} & 
        		 \includegraphics[width=1.8cm,height=1.8cm]{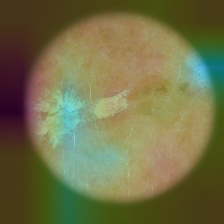} & 
        		 \includegraphics[width=1.8cm,height=1.8cm]{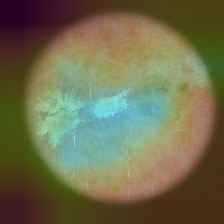} \\
        		
        		\rotatebox{90}{\textcolor{white}{Lg}}
        		 \includegraphics[width=1.8cm,height=1.8cm]{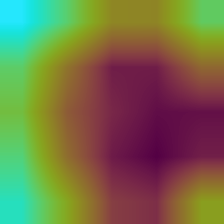} & 
        		 \includegraphics[width=1.8cm,height=1.8cm]{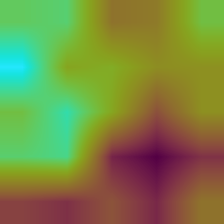} & 
        		 \includegraphics[width=1.8cm,height=1.8cm]{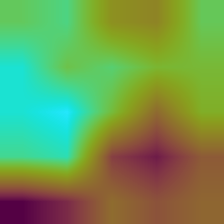} & 
        		 \includegraphics[width=1.8cm,height=1.8cm]{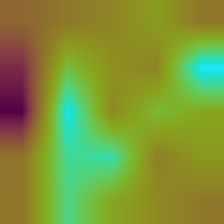} & 
        		 \includegraphics[width=1.8cm,height=1.8cm]{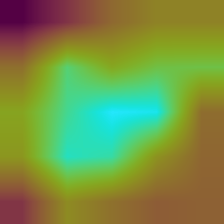} \\
        		
        		\rotatebox{90}{L-DCA \textcolor{white}{Lg}}
        		 \includegraphics[width=1.8cm,height=1.8cm]{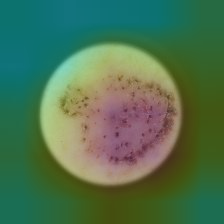} & 
        		 \includegraphics[width=1.8cm,height=1.8cm]{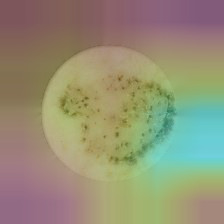} & 
        		 \includegraphics[width=1.8cm,height=1.8cm]{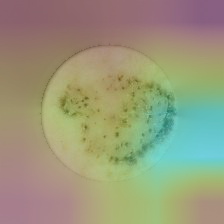} & 
        		 \includegraphics[width=1.8cm,height=1.8cm]{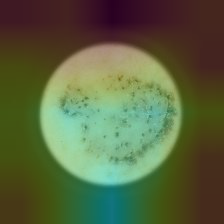} & 
        		 \includegraphics[width=1.8cm,height=1.8cm]{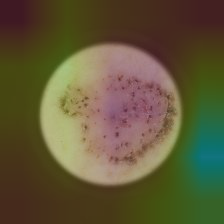} \\
        		
        		\rotatebox{90}{\textcolor{white}{Lg}}
        		 \includegraphics[width=1.8cm,height=1.8cm]{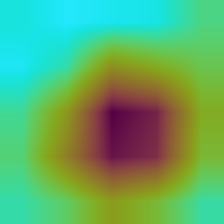} & 
        		 \includegraphics[width=1.8cm,height=1.8cm]{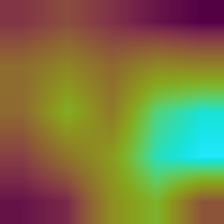} & 
        		 \includegraphics[width=1.8cm,height=1.8cm]{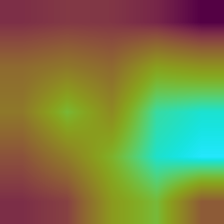} & 
        		 \includegraphics[width=1.8cm,height=1.8cm]{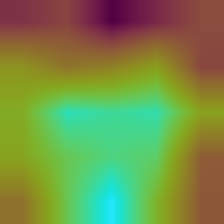} & 
        		 \includegraphics[width=1.8cm,height=1.8cm]{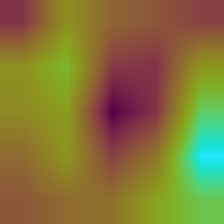} \\
        		
        		\rotatebox{90}{O-DCA \textcolor{white}{Lg}}
        		 \includegraphics[width=1.8cm,height=1.8cm]{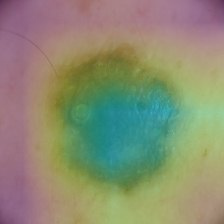} & 
        		 \includegraphics[width=1.8cm,height=1.8cm]{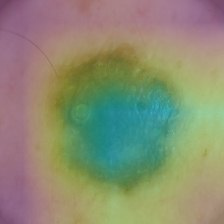} & 
        		 \includegraphics[width=1.8cm,height=1.8cm]{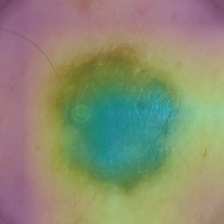} & 
        		 \includegraphics[width=1.8cm,height=1.8cm]{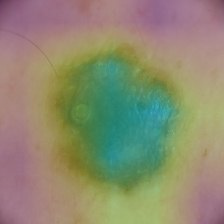} & 
        		 \includegraphics[width=1.8cm,height=1.8cm]{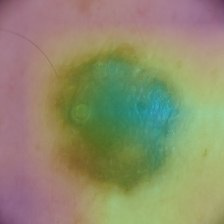} \\
        		
        		\rotatebox{90}{\textcolor{white}{Lg}}
        		 \includegraphics[width=1.8cm,height=1.8cm]{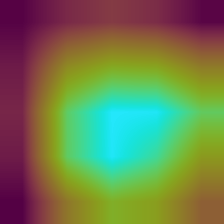} & 
        		 \includegraphics[width=1.8cm,height=1.8cm]{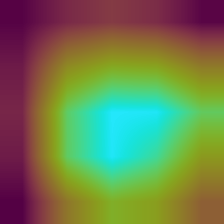} & 
        		 \includegraphics[width=1.8cm,height=1.8cm]{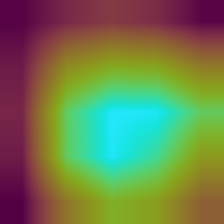} & 
        		 \includegraphics[width=1.8cm,height=1.8cm]{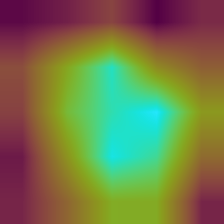} & 
        		 \includegraphics[width=1.8cm,height=1.8cm]{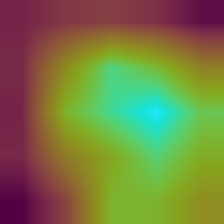} \\
        
        	\end{tabular}
        	\caption[]{Grad-CAM heatmap results to visually compare activations of different models on the test set with small (S-DCA), medium (M-DCA), large (L-DCA), and other (O-DCA) DCA sizes. Clean Model (CM) is the model (clean images) used for classification on the test set (with DCA); Navier-Stokes (NS) is the model used for classification on the test set inpainted by Navier-Stokes; Telea (Tel) is the model used for classification on the test set inpainted by Telea; Binary (Bin) is the model (clean images superimposed with synthetic binary DCAs) used for classification on the test set (with DCA); and Realistic (Real) is the model (clean images superimposed with synthetic realistics DCAs) used for classification on the test set (with DCA).}
        	\label{fig:fig08}
        \end{figure*}

To further confirm and compare the differences between the methods, Figure~\ref{fig:fig08} shows the output of Grad-CAM heatmap activations across the different networks. Each column in this figure shows a different method used for evaluation, and each pair of rows in the figure represent the testing set and the activations shown with Grad-CAM.

        
Several observations are noted in Figure~\ref{fig:fig08}. Without any bias, we selected the images randomly from each DCA size. Firstly, we can see that when comparing the performance on small DCA, every method improved the focus of the network and expanded the focus from the upper left corner to the rest of the image. It is difficult to determine the most efficient method from the results generated for this particular example of small DCA as none of the results show a perfect focus area. When analysing the performance on medium DCA, it can be seen that all evaluation methods showed a large improvement on the original network activations. For each of the methods the focus is taken away from the DCA and distributed across the image. The method showing the best fitting activations to what are expected can be seen in the results from the model trained on realistic DCA.
        
When comparing the activations for the large DCA test sets, a scenario similar to the medium DCA results is observed where the activations are mostly focused on the DCA region on the baseline results with the clean model, whereas most other methods manage to disrupt this focus and focus more on the areas of interest. 
The method displaying the best overall focus is the model trained on binary DCA, where the activations focus almost entirely on the central region of the image. The results for the DCA consuming less than 1\% of the image do not display much variability compared to the baseline generated on the clean model with original data. This is due to the minimal area that the DCA occupies.
        
\subsection{DCA removal combined with synthetic training}
    
When evaluating the networks across each of the testing sets, the images inpainted with both Navier-Stokes and Telea methods were also used to evaluate both of the models trained on synthetic DCA images. Table~\ref{tab:tab06} shows the results generated from evaluating the Binary DCA and Realistic DCA with images inpainted from both methods.
        
        \begin{table*}[!htbp]
            \centering
            \caption{Performance evaluation of the superimposed synthetic DCA training models on DCA inpainted testing sets. It is noted that the superimposed binary DCA model performed better in TPR, but the superimposed realistic DCA model performed better in TNR on the original test set than the inpainted images.}
            \scalebox{0.85}{
            \begin{tabular}{llllllll}
                \hline
                \textbf{Model Used} & \textbf{Test Set} & \multicolumn{6}{l}{\textbf{Metrics}} \\
                \cline{3-8}
                 & & \textbf{Acc} & \textbf{TPR} & \textbf{TNR} & \textbf{Precision} & \textbf{F1} & \textbf{AUC} \\
                \hline
                \hline
                Binary DCA & Original - small & 0.61 & \textbf{0.90} & 0.33& 0.57 & 0.70 & 0.67 \\
                 & NS - small & 0.61 & 0.89 & 0.33  & 0.57& 0.70 & 0.67\\
                 & Telea - small & 0.61 & 0.89 & 0.34  & 0.57& 0.70 & 0.67\\
                \cline{2-8}
                 & Original - medium & 0.63 & \textbf{0.94} & 0.31 & 0.58& 0.72 & 0.68 \\
                 & NS - medium & 0.63 & 0.91 & 0.36 & 0.59& 0.71 & 0.70 \\
                 & Telea - medium & 0.64 & 0.89 & 0.40 & 0.60& 0.71 & 0.71 \\
                \cline{2-8}
                 & Original - large & 0.55 & \textbf{0.96} & 0.13 & 0.53& 0.68 & 0.62 \\
                 & NS - large & 0.65 & 0.73 & 0.56 & 0.62& 0.67 & 0.69 \\
                 & Telea - large & 0.64 & 0.65 & 0.62 & 0.64& 0.64 & 0.71 \\
                \cline{2-8}
                 & Original - oth & 0.60 & \textbf{0.83} & 0.36 & 0.57& 0.67 & 0.67 \\
                 & NS - oth & 0.59 & 0.81 & 0.38 & 0.56& 0.67 & 0.66 \\
                 & Telea - oth & 0.59 & 0.81 & 0.38 & 0.56& 0.67 & 0.66 \\
                \hline 
                \hline 
                
                Realistic DCA & Original - small & 0.60 & 0.85 & 0.35& 0.57 & 0.68 & 0.65 \\
                 & NS - small & 0.60 & 0.85 & 0.34 & 0.56& 0.68 & 0.65 \\
                 & Telea - small & 0.60 & 0.85 & 0.35 & 0.57& 0.68 & 0.65 \\
                \cline{2-8}
                 & Original - medium & 0.64 & 0.75 & \textbf{0.53} & \textbf{0.62}& 0.68 & 0.70 \\
                 & NS - medium & 0.63 & 0.87 & 0.39 & 0.59& 0.70 & 0.67 \\
                 & Telea - medium & 0.63 & 0.84 & 0.43 & 0.60& 0.70 & 0.68 \\
                \cline{2-8}
                 & Original - large & 0.60 & 0.39 & \textbf{0.80}& \textbf{0.66} & 0.49 & 0.63 \\
                 & NS - large & 0.59 & 0.60 & 0.58 & 0.59& 0.60 & 0.64 \\
                 & Telea - large & 0.60 & 0.49 & 0.70 & 0.62& 0.55 & 0.64 \\
                \cline{2-8}
                 & Original - oth & 0.58 & 0.81 & 0.35 & 0.55& 0.66 & 0.65 \\
                 & NS - oth & 0.57 & 0.79 & 0.36 & 0.55& 0.65 & 0.64 \\
                 & Telea - oth & 0.57 & 0.79 & 0.36 & 0.55& 0.65 & 0.64 \\

                \hline
                \hline
            \end{tabular}
            }
            \label{tab:tab06}
        \end{table*}

As can be seen in Table~\ref{tab:tab06}, the accuracy for superimposed binary DCA increases but the accuracy for superimposed realistics DCA decreases when compared original test sets with inpainted test sets. For medium DCA test sets, the realistic DCA model achieved the best result on original test set. In contrast, the binary model achieved better results when evaluated using inpainted images. The largest discrepancy is exhibited in the large DCA test set, where the binary DCA model achieved the best result in TPR of 0.96, but the poorest result in TNR of 0.13. When evaluating with inpainted images, a large increase in TNR for binary DCA model can be seen.

        \begin{figure}[!h]
        	\centering
        	\begin{tabular}{cccc}
        	
        	     Binary Base & Clean NS & Binary NS & Realistic NS \\

        		 \includegraphics[width=1.8cm,height=1.8cm]{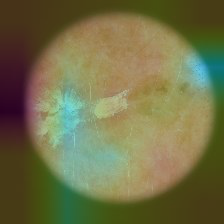} & 
        		 \includegraphics[width=1.8cm,height=1.8cm]{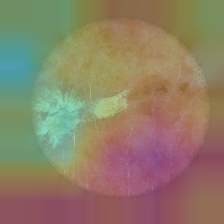} & 
        		 \includegraphics[width=1.8cm,height=1.8cm]{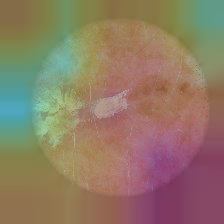} & 
        		 \includegraphics[width=1.8cm,height=1.8cm]{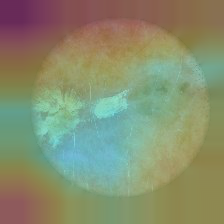} \\
        		 
        		 \includegraphics[width=1.8cm,height=1.8cm]{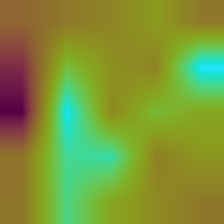} & 
        		 \includegraphics[width=1.8cm,height=1.8cm]{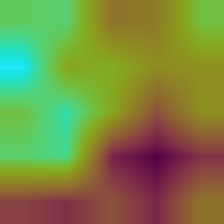} & 
        		 \includegraphics[width=1.8cm,height=1.8cm]{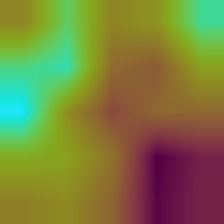} & 
        		 \includegraphics[width=1.8cm,height=1.8cm]{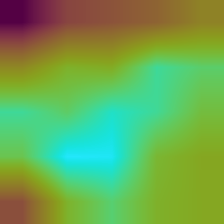} \\
        		 
        		 Binary Base & Clean Telea & Binary Tel & Realistic Tel \\

        		 \includegraphics[width=1.8cm,height=1.8cm]{fig14_binary_base_overlay.png} & 
        		 \includegraphics[width=1.8cm,height=1.8cm]{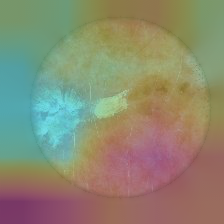} & 
        		 \includegraphics[width=1.8cm,height=1.8cm]{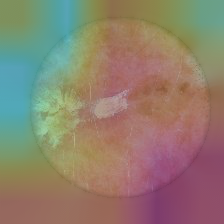} & 
        		 \includegraphics[width=1.8cm,height=1.8cm]{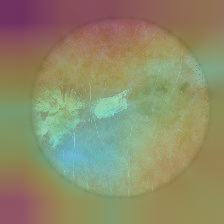} \\
        		 
        		 \includegraphics[width=1.8cm,height=1.8cm]{fig14_binary_base_heatmap.png} & 
        		 \includegraphics[width=1.8cm,height=1.8cm]{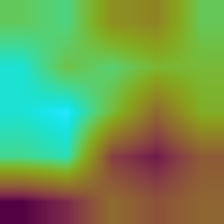} & 
        		 \includegraphics[width=1.8cm,height=1.8cm]{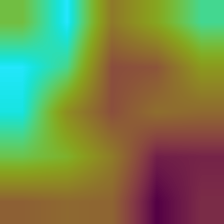} & 
        		 \includegraphics[width=1.8cm,height=1.8cm]{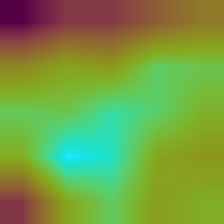} \\
        	\end{tabular}
        	\caption[]{Grad-CAM results for medium inpainted images on networks trained with synthetic DCA. Tel - Telea.}
        	\label{fig:fig14}
        \end{figure}

\begin{figure}[!htbp]
        	\centering
        	\begin{tabular}{cccc}
        	
        	     Binary Base & Clean NS & Binary NS & Realistic NS \\

        		 \includegraphics[width=1.8cm,height=1.8cm]{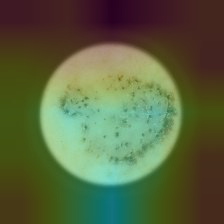} & 
        		 \includegraphics[width=1.8cm,height=1.8cm]{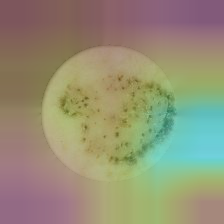} & 
        		 \includegraphics[width=1.8cm,height=1.8cm]{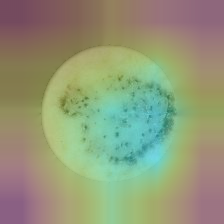} & 
        		 \includegraphics[width=1.8cm,height=1.8cm]{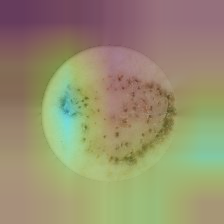} \\
        		 
        		 \includegraphics[width=1.8cm,height=1.8cm]{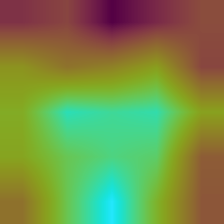} & 
        		 \includegraphics[width=1.8cm,height=1.8cm]{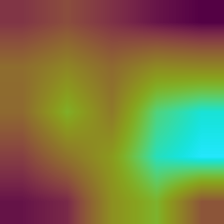} & 
        		 \includegraphics[width=1.8cm,height=1.8cm]{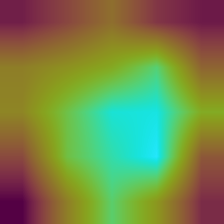} & 
        		 \includegraphics[width=1.8cm,height=1.8cm]{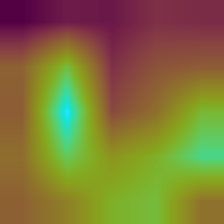} \\
        		 
        		 Binary Base & Clean Telea & Binary Tel & Realistic Tel \\

        		 \includegraphics[width=1.8cm,height=1.8cm]{fig15_binary_base_overlay.png} & 
        		 \includegraphics[width=1.8cm,height=1.8cm]{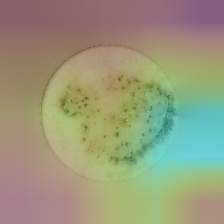} & 
        		 \includegraphics[width=1.8cm,height=1.8cm]{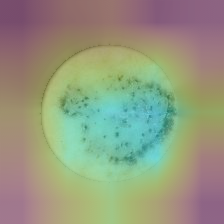} & 
        		 \includegraphics[width=1.8cm,height=1.8cm]{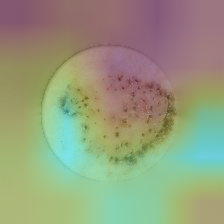} \\
        		 
        		 \includegraphics[width=1.8cm,height=1.8cm]{fig15_binary_base_heatmap.png} & 
        		 \includegraphics[width=1.8cm,height=1.8cm]{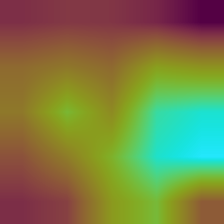} & 
        		 \includegraphics[width=1.8cm,height=1.8cm]{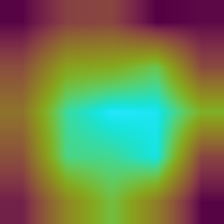} & 
        		 \includegraphics[width=1.8cm,height=1.8cm]{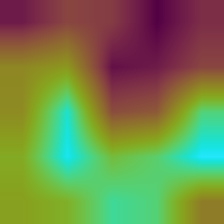} \\
        	\end{tabular}
        	\caption[]{Grad-CAM results for large inpainted images on networks trained with synthetic DCA. Tel - Telea.}
        	\label{fig:fig15}
        \end{figure}

Figure~\ref{fig:fig14} shows the different heatmaps generated for an image containing a medium DCA and Figure~\ref{fig:fig15} shows the different heatmaps generated for an image containing a large DCA. The same image is inpainted and examined again in each network to determine the differences in activations. The activations for the `small' and `other' DCA sizes do not show significant differences. To aid the comparison between the base test set and the inpainted test sets, the model trained on binary DCA using the baseline test set is used as it produces the best TPR across each of the DCA sizes present as seen in Table~\ref{tab:tab06}.


It can be seen that the activations for both inpainting methods produce similar results for each of the DCA images sizes. For the medium Binary Base, Clean, and Binary NS/Telea images, the activations are generalised around both lesion and DCA regions, while the medium Realistic activations show significantly more concentrated focus towards the lesion region. For large Binary Base, Clean, and Binary NS/Telea, the activations are generally more focused on the lesion regions, with the large Realistic images showing a much more generalised spread of activations across lesion and DCA. Whilst the overall performance of the networks appear to improve when inpainted images are evaluated on a network trained by images containing synthetic DCA, the class activation maps show that the area in which the network is focused is not as targeted on the lesion regions as the intended test sets.

         
\subsection{Heatmap Contrast and Brightness Intensity Measures}

Table~\ref{tab:tab07} shows a breakdown of heatmap contrast and brightness intensity measures for all images for each DCA size and all models. We observe from the heatmap intensity results that almost all models show a clear bias towards focussing on internal lesion regions, with the exception of the clean original small, medium, and large models which show a clear bias of network focus towards the external DCA regions (shown in the negative mean difference values). The clean original small model shows a slight bias towards external DCA regions (RMS mean diff = -2.50, avg. brightness mean diff = -0.53), while the clean original medium model (RMS mean diff = -40.28, avg. brightness mean diff = -40.57) and the clean original large model (RMS mean diff = -44.87, avg. brightness mean diff = -48.44) show large biases towards external DCA regions ($>$ -40).

\begin{table*}[!h]
            \centering
            \caption{Contrast and brightness intensity measures according to DCA size. RMS - root mean square; mean difference = $internal\;mean - external\;mean$ (highest value shows a higher ratio of activations in the target area; positive for lesion area and negative for DCA area); Std - standard deviation.}
            \scalebox{0.53}{
            \begin{tabular}{llllllllllll}
                \hline
                \textbf{Model Used} & \textbf{Test Set} & \multicolumn{5}{l}{\textbf{RMS}} & \multicolumn{5}{l}{\textbf{Avg. Brightness}} \\
                \cline{3-12}
                && \multicolumn{2}{l}{\textbf{Internal}} & \multicolumn{2}{l}{\textbf{External}} & & \multicolumn{2}{l}{\textbf{Internal}} & \multicolumn{2}{l}{\textbf{External}} & \\
                \cline{3-12}
                && Mean $\Uparrow$ & Std $\Downarrow$ & Mean $\Downarrow$ & Std $\Downarrow$ & Diff (Mean) & Mean $\Uparrow$ & Std $\Downarrow$ & Mean $\Downarrow$ & Std $\Downarrow$ & Diff (Mean)\\
                \hline
                \hline
                Clean & Original - small & 111.10 & 22.85 & 113.60 & \textbf{23.10} &   -2.50      & 107.43 & 23.21 & 107.96 & \textbf{23.57} & -0.53\\
                & NS - small & 125.07 & 14.22 & 107.11 & 27.06 &    17.96         & 121.25 & 15.60 & 102.14 & 27.51 & 19.11\\
                & Telea - small & \textbf{125.12} & \textbf{14.13} & \textbf{107.05} & 27.18 & \textbf{18.07}            & \textbf{121.28} & \textbf{15.51} & \textbf{102.12} & 27.64 & \textbf{19.16}\\
                \cline{2-12}
                & Original - medium & 89.82 & \textbf{14.24} & 130.10 & \textbf{7.89} &      -40.28      & 85.04 & \textbf{14.81} & 125.61 & \textbf{9.00} & -40.57 \\
                & NS - medium & 123.69 & 15.42 & 115.37 & 19.44 & 8.32            & 120.35 & 16.57 & 110.86 & 20.19 & 9.49\\
                & Telea - medium & \textbf{125.10} & 14.97 & \textbf{115.06} & 19.94 &    \textbf{10.04}         & \textbf{121.84} & 16.15 & \textbf{110.58} & 20.70 & \textbf{11.26}\\
                \cline{2-12}
                & Original - large & 103.59 & \textbf{10.07} & 148.46 & \textbf{6.89} &      -44.87      & 98.51 & \textbf{10.65} & 146.95 & \textbf{7.82} & -48.44\\
                & NS - large & 132.14 & 17.52 & \textbf{115.95} & 17.23 &    \textbf{16.19}         & 130.03 & 18.69 & \textbf{111.67} & 18.07 & \textbf{18.36}\\
                & Telea - large & \textbf{132.26} & 17.18 & 116.78 & 17.69 & 15.48            & \textbf{130.14} & 18.30 & 112.53 & 18.62 & 17.61\\
                \cline{2-12}
                & Original - oth & 124.23 & 14.34 & 99.18 & 28.46 &     25.05      & 120.18 & \textbf{15.53} & 93.93 & 28.67 & 26.25\\
                & NS - oth & \textbf{124.84} & 14.31 & 98.88 & \textbf{28.18} & 25.96            & \textbf{120.82} & 15.57 & 93.64 & \textbf{28.36} & 27.18\\
                & Telea - oth & 124.81 & \textbf{14.30} & \textbf{98.79} & 28.19 &    \textbf{26.02}         & 120.78 & 15.56 & \textbf{93.55} & 28.39 & \textbf{27.23}\\
                \hline
                \hline
                Binary DCA & Original - small & \textbf{132.79} & 11.00 & \textbf{111.27} & \textbf{23.86} &     \textbf{21.52}     & \textbf{129.88} & 11.93 & \textbf{106.24} & \textbf{24.92} & \textbf{23.64}\\
                & NS - small & 132.40 & 10.84 & 112.95 & 24.74 &    19.45         & 129.43 & 11.71 & 108.13 & 25.93 & 21.30\\
                & Telea - small & 132.23 & \textbf{10.83} & 113.14 & 25.02 & 19.09            & 129.23 & \textbf{11.69} & 108.40 & 26.25 & 20.83\\
                \cline{2-12}
                & Original - medium & \textbf{135.98} & \textbf{10.67} & \textbf{111.41} & \textbf{13.70} &      \textbf{24.57}     & \textbf{133.84} & \textbf{11.52} & \textbf{106.79} & \textbf{14.07} & \textbf{27.05}\\
                & NS - medium & 134.52 & 11.01 & 117.09 & 16.56 & 17.43            & 132.20 & 11.86 & 112.85 & 17.34 & 19.35\\
                & Telea - medium & 134.30 & 10.88 & 118.40 & 17.54 &    15.90         & 131.94 & 11.73 & 114.27 & 18.50 & 17.67\\
                \cline{2-12}
                & Original - large & \textbf{136.85} & \textbf{12.71} & \textbf{114.98} & \textbf{14.08} &      \textbf{21.87}      & \textbf{135.22} & \textbf{13.46} & \textbf{110.93} & \textbf{14.64} & \textbf{24.29}\\
                & NS - large & 135.34 & 15.42 & 120.65 & 17.64 &    14.69         & 133.53 & 16.32 & 116.98 & 18.46 & 16.55\\
                & Telea - large & 133.78 & 16.08 & 123.18 & 18.69 & 10.60            & 131.85 & 17.06 & 119.70 & 19.70 & 12.15\\
                \cline{2-12}
                & Original - oth & 130.74 & \textbf{10.41} & \textbf{109.04} & \textbf{32.06} &     21.70      & 127.27 & \textbf{11.03} & \textbf{104.42} & \textbf{33.83} & 22.85\\
                & NS - oth & \textbf{130.82} & 10.89 & 109.06 & 32.16 & \textbf{21.76}            & \textbf{127.37} & 11.52 & 104.46 & 33.95 & \textbf{22.91}\\
                & Telea - oth & 130.81 & 10.97 & 109.06 & 32.23 & 21.75            & 127.34 & 11.61 & 104.47 & 34.02 & 22.87\\
                \hline
                \hline
                Realistic DCA & Original - small & \textbf{130.83} & 11.24 & \textbf{111.81} & \textbf{24.62} &      \textbf{19.02}       & \textbf{127.66} & 12.28 & \textbf{106.73} & \textbf{25.38} & \textbf{20.93}\\
                & NS - small & 130.17 & 11.16 & 112.31 & 25.38 &    17.86         & 126.93 & 12.14 & 107.47 & 26.28 & 19.46\\
                & Telea - small & 130.23 & \textbf{11.11} & 112.20 & 25.63 & 18.03            & 126.98 & \textbf{12.10} & 107.35 & 26.55 & 19.63\\
                \cline{2-12}
                & Original - medium & 130.95 & 12.29 & \textbf{118.04} & 16.91 &       12.91      & 128.32 & 13.26 & \textbf{113.74} & 17.61 & 14.58\\
                & NS - medium & 131.61 & 12.57 & 118.56 & \textbf{16.27} &      13.05        & 129.19 & 13.46 & 114.12 & \textbf{16.90} & 15.07\\
                & Telea - medium & \textbf{132.58} & \textbf{11.92} & 118.89 & 16.91 &    \textbf{13.69}         & \textbf{130.17} & \textbf{12.81} & 114.48 & 17.64 & \textbf{15.69}\\
                \cline{2-12}
                & Original - large & 130.29 & 17.74 & 123.03 & 14.72 &      7.26       & 128.26 & 18.83 & 119.59 & \textbf{15.54} & 8.67\\
                & NS - large & \textbf{136.25} & \textbf{15.52} & \textbf{118.74} & \textbf{13.16} &    \textbf{17.51}         & \textbf{134.54} & \textbf{16.54} & \textbf{114.66} & 15.78 & \textbf{19.88}\\
                & Telea - large & 135.18 & 15.53 & 120.55 & 16.21 & 14.63            & 133.31 & 16.59 & 116.68 & 17.06 & 16.63\\
                \cline{2-12}
                & Original - oth & 129.19 & \textbf{10.81} & 108.41 & 27.92 &      20.78     & 125.67 & \textbf{11.51} & 103.62 & 29.27 & 22.05\\
                & NS - oth & 129.42 & 11.50 & \textbf{108.20} & \textbf{27.89} & \textbf{21.22}            & 125.90 & 12.23 & \textbf{103.41} & \textbf{29.25} & \textbf{22.49}\\
                & Telea - oth & \textbf{129.47} & 11.53 & 108.34 & 27.97 & 21.13            & \textbf{125.95} & 12.28 & 103.56 & 29.35 & 22.39\\
                \hline

                \hline
            \end{tabular}
            }
            \label{tab:tab07}
        \end{table*}

The model with the highest RMS contrast and highest average brightness intensity bias towards internal (lesion) regions is the binary DCA original large model (RMS mean = 136.85, avg brightness mean = 135.22). These results correlate with the TPR for this model (Table \ref{tab:tab06}) which has the highest TPR (0.96). However, the correlation is not present for accuracy, TNR, precision, and AUC, which were shown to be the lowest reported metrics for this model. This is likely to be a consequence of the binary DCA original large model comprising of mostly melanoma examples, which makes TPR occurrences more prevalent.

The model with the highest RMS mean and highest average brightness mean for external DCA regions is the clean original large model (RMS mean = 148.46, avg brightness mean = 146.95). These results correlate with the TPR (0.39) and F1-score (0.49) results for this model (Table \ref{tab:tab06}), which shows the lowest results for these metrics, indicating that the model was focussed more on external DCA regions.

The model with the highest RMS mean difference and highest average brightness mean difference is the clean Telea `other' model (RMS mean diff = 26.02, avg brightness mean diff = 27.23). The intensity metrics for this model indicate that it showed the highest shift in network focus towards internal (lesion) features. However, this shift alone was not sufficient enough to provide it with the highest overall scores in accuracy, TPR, TNR, precision, F1-score, and AUC.

\section{Discussion}

Between Figure~\ref{fig:fig02}, Figure~\ref{fig:fig03}, Figure~\ref{fig:fig04}, and Figure~\ref{fig:fig05} the Grad-CAM heatmap activations show relatively similar results across each of the DCA sizes fed into the network. For the small, medium and large DCA, improvements in the focus of the activations can be seen for both of the models trained on binary and realistic DCA. Both models exhibit similar results. Results for DCA covering less than 1\% of the image show similar or worse activations on models trained with DCA images when comparing to the clean model.

We observe that the binary DCA model achieved better TPR than the realistic DCA model in our experiments, but it did not outperform the clean model. A possible reason for this could be that the gradient inherent in the smooth transitions between skin and DCA may represent an introduction of further complex features that creates an additional learning challenge to the network. However, with the use of superimposed realistic DCA, a notable improvement in TNR and precision indicate the network was able to learn to handle the DCA and was capable of reducing the biases of classifying DCA as melanoma. 

\begin{figure*}[!htbp]
	\centering
	\begin{tabular}{ccc}
	
	    \rotatebox{90}{True Positive \textcolor{white}{Lg}}
		\includegraphics[width=3.4cm,height=3.4cm]{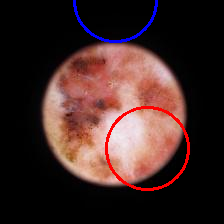} &
		\includegraphics[width=3.4cm,height=3.4cm]{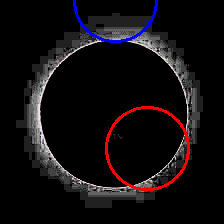} &
		\includegraphics[width=3.4cm,height=3.4cm]{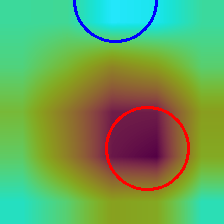} \\
		
		\rotatebox{90}{False Positive \textcolor{white}{Lg}}
		\includegraphics[width=3.4cm,height=3.4cm]{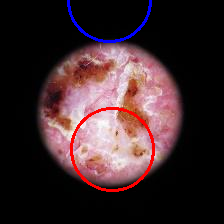} &
		\includegraphics[width=3.4cm,height=3.4cm]{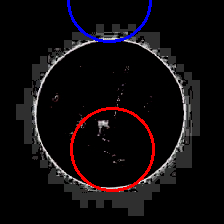} &
		\includegraphics[width=3.4cm,height=3.4cm]{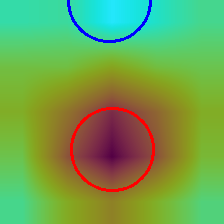} \\
		
		\rotatebox{90}{True Negative}
		\includegraphics[width=3.4cm,height=3.4cm]{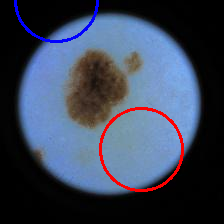} &
		\includegraphics[width=3.4cm,height=3.4cm]{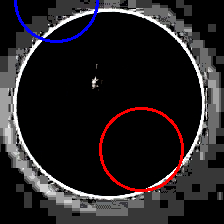} &
		\includegraphics[width=3.4cm,height=3.4cm]{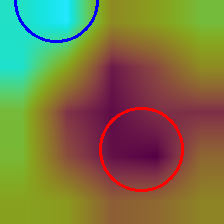} \\
		
		\rotatebox{90}{False Negative}
		\includegraphics[width=3.4cm,height=3.4cm]{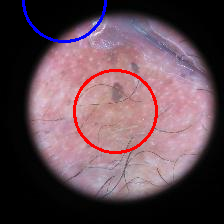} &
		\includegraphics[width=3.4cm,height=3.4cm]{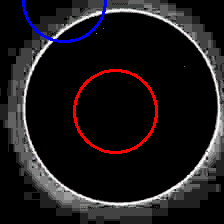} &
		\includegraphics[width=3.4cm,height=3.4cm]{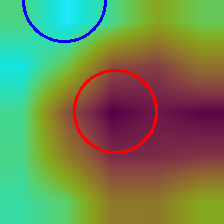} \\
		
		Original Image & Increased Contrast & Prediction Heatmap \\
	\end{tabular}
	\caption[]{Illustration of predictions on test images from the clean large DCA model. Images shown are the original unaltered test images (1st column), original images with increased contrast to expose light leakage (2nd column), and the corresponding prediction heatmaps (3rd column). The first row shows true positives, the second row shows true negatives, and the third column shows false negatives. Blue circles indicate the brightest region of the heatmap, red circles indicate the darkest region of the heatmap. Brightest and darkest heatmap regions were obtained using the minMaxLoc function in the OpenCV library (\cite{bradski2000opencv}).}
	\label{fig:leakage}
\end{figure*}

During our analysis of the heatmaps for the skin lesion images, we observed that many of the heatmaps showed that networks would focus on DCA regions. We hypothesised that the DCA regions may exhibit artifacts that were not visible to the human eye and were causing networks to focus on these areas. For example, we speculated that there may be JPG artifacts present within the DCA areas that introduced additional complex features into the DCA regions, or that the DCA region did not comprise of only black pixels. To test for this, we performed an additional analysis on images taken from the clean large DCA model. Images were selected from the test results of this model as it exhibited a heavy bias towards classifying almost all test images as melanoma. We adjusted the contrast of the original lesion images which naturally exhibit DCA. This process revealed that many lesion images with DCA exhibited complex pixel patterns that radiate outwards from the border regions between lesion area and DCA area that would not ordinarily be visible to the human eye. Due to the uniformity of these pixel patterns around the lesion / DCA borders, we speculate that these patterns are the result of light leakage from the dermoscope. Many dermoscope models are equipped with a built-in array of LED lights that surround the perimeter of the lens. Figure \ref{fig:leakage} shows test images together with corresponding increased contrast images and Grad-CAM heatmap activation images from the clean large DCA model. The extent to which the light leakage is present can vary between examples. This may be due to the use of different dermoscope models (size and quality), dermoscope settings, or the amount of pressure applied to the skin during an examination. If a dermoscope is powered by a battery, the amount of light leakage may also be affected by battery power levels which would affect the intensity of light being emitted by the device.

\begin{figure*}[!htbp]
	\centering
	\begin{tabular}{cccc}
		\includegraphics[width=2.5cm,height=2.5cm]{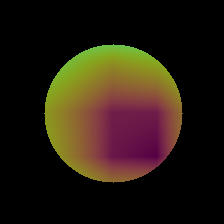} &
		\includegraphics[width=2.5cm,height=2.5cm]{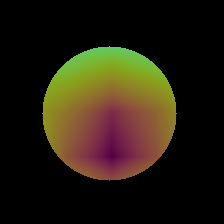} &
		\includegraphics[width=2.5cm,height=2.5cm]{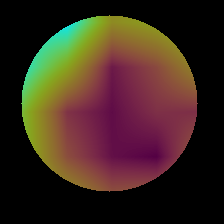} &
		\includegraphics[width=2.5cm,height=2.5cm]{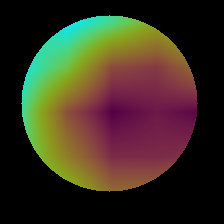} \\
		True Positive & False Positive & True Negative & False Negative \\
	\end{tabular}
	\caption[]{Illustration of Grad-CAM heatmaps with the original generated masks overlaid to show network activation levels within lesion regions. Heatsmaps were taken from inference results for the clean large DCA model.}
	\label{fig:leakage_masks}
\end{figure*}

We observe from the images in Figure \ref{fig:leakage} that most examples show that the network focussed mainly on the surrounding DCA regions, including areas of light leakage, regardless of the prediction result. We draw two conclusions from this: (1) the network would use outer regions of the image to form both correct and incorrect predictions that may be due to the presence of black pixels, the complex features introduced by the light leakage, or a combination of both, and (2) the network may be learning spurious correlations between the features in the DCA and light leakage areas within the DCA, hence the apparent randomness of the results. As shown in Figure \ref{fig:leakage}, we indicate the brightest (blue circles) and darkest (red circles) regions of the heatmap activation images, with bright regions representing the highest levels of network focus, and dark regions indicating the lowest levels of network focus. In these examples, lesion details are present in the outer regions of the lesion area which are positioned close to the DCA regions. It may therefore be possible that although the network appears to be focusing mostly on DCA regions, for correct predictions the network is able to determine class using those lesion features that are close to the DCA perimeter. Figure \ref{fig:leakage_masks} shows the heatmaps from Figure \ref{fig:leakage} with masks applied to show that the network still directs some of its focus towards the actual lesion regions.


The DCA and light leakage results shown in Figure \ref{fig:leakage} clearly illustrate that the clean large DCA model is still prone to focusing on pure black DCA regions and light leakage regions to make correct and incorrect classification predictions. These results also indicate that the shift in activations causes the model to focus significantly less on the actual lesion regions. In the examples shown, the areas with the most focus all have light leakage artifacts present to varying degrees and varying levels of visual complexity.

\section{Conclusion}
While existing research attempted to remove artifacts and focused on creating improved deep learning models for melanoma classification, we emphasise on better understanding of the data and behaviour of the learning process, which are the keys to provide new insights into skin lesion analysis. Existing research shows that the limited work focusing on DCA is mostly inconclusive, mainly due to a lack of publicly available DCA datasets to support the task. Therefore, we introduce a new curated balanced dataset with an equal number of melanoma and non-melanoma cases, drawn from publicly available skin lesion image datasets, which consists of 6126 training images without DCA and 4124 test images with DCA.

We investigated the effect of DCA in dermoscopic images in melanoma classifcation by producing a baseline result using the proposed training and test sets. As expected, we achieved high TPR and poor TNR, this is due to the tendency of the model to classify DCA as melanoma. We demonstrated that DCA removal and inpainting methods improved the results marginally and proposed a new strategy to address the negative effect of DCAs, i.e., superimposed synthetic DCAs in the training set to train the deep learning model. In addition to existing Binary DCAs, we developed a new synthetic DCA method (namely, Realistic DCA) to improve the realism of the DCA appearance when compared to naturally occurring DCA. We present results from experiments performed on these two artificially generated DCA types and demonstrate their effect in comparison to inpainting of real DCA. Our results indicate that binary DCA provided the highest TPR but realistic DCA provided the highest TNR. Our experiments showed that the removal and inpainting of DCAs is not the sole solution to improve the performance of deep learning models. Instead, our experiments using superimposed synthetic DCAs improved the TNR and precision of melanoma classification. We recommend further investigation to focus on superimposed DCA rather than DCA removal and inpainting as the latter is computationally expensive, achieved marginal improvement, and it is not clear of what new element was introduced in the inpainting process. Moreover, a notable improvement in TNR and precision when using superimposed synthetic DCAs provide some early indication of the capability of such a proposal to reduce the biases of classifying DCA as melanoma.

We interpreted the performance of the deep learning model on different settings by using Grad-CAM heatmap visualisation and its association with the dermoscopy light leakage. We observed that the DCA regions may exhibit artifacts that were not visible to the human eye where the deep learning model might tend to use those features for decision making. Another interesting observation is that although the focus is on the region of interest (skin lesions), there is an apparent randomness in the predictions due to the challenging nature of melanoma classification.

We developed a new quantitative method based on heatmap contrast and brightness intensity measures to increase the understanding of the differences between internal and external DCA regions. Our method for quantifying Grad-CAM heatmap activation images shows a good correlation between heatmap contrast and brightness intensity and recorded metrics such as accuracy and F1-score. This measure can be used in other research domains where heatmap visualisation is used. All relevant source code and guidelines to obtain the dataset will be made available upon acceptance of the paper. 

\backmatter

\bibliography{sn-bibliography}


\end{document}